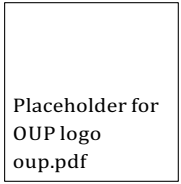
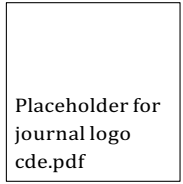

Placeholder for OUP logo oup.pdf

Placeholder for journal logo cde.pdf

RESEARCH ARTICLE

# A new approach for solving global optimization and engineering problems based on modified Sea Horse Optimizer

Fatma A. Hashim[1], Reham R. Mostafa[2,3], Ruba Abu Khurma[4], Raneem Qaddoura[5] and P.A. Castillo[6,*]

[1]Faculty of Engineering, Helwan University, Egypt. Email: fatma_hashim@h-eng.helwan.edu.eg and [2]Research Institute of Sciences and Engineering (RISE), University of Sharjah, Sharjah 27272, United Arab Emirates. Email: REldeiasti@sharjah.ac.ae and [3]Department of Information System, Faculty of Computers and Information Sciences, Mansoura University, Mansoura 35516, Egypt. Email: reham_2006@mans.edu.eg and [4]MEU Research Unit, Faculty of Information Technology, Middle East University, Amman, 11831, Jordan. Email: ruba_abukhurma@yahoo.edu.jo and [5]School of Computing and Informatics, Al Hussein Technical University, Amman, Jordan. Email:raneem.qaddoura@htu.edu.jo and [6]Department of Computer Engineering, Automatics and Robotics, University of Granada, Granada, Spain. Email: pacv@ugr.es

*Corresponding author. E-mail: Email: pacv@ugr.es

## Abstract

Sea Horse Optimizer (SHO) is a noteworthy metaheuristic algorithm that emulates various intelligent behaviors exhibited by sea horses, encompassing feeding patterns, male reproductive strategies, and intricate movement patterns. To mimic the nuanced locomotion of sea horses, SHO integrates the logarithmic helical equation and Levy flight, effectively incorporating both random movements with substantial step sizes and refined local exploitation. Additionally, the utilization of Brownian motion facilitates a more comprehensive exploration of the search space. This study introduces a robust and high-performance variant of the SHO algorithm named mSHO. The enhancement primarily focuses on bolstering SHO's exploitation capabilities by replacing its original method with an innovative local search strategy encompassing three distinct steps: a neighborhood-based local search, a global non-neighbor-based search, and a method involving circumnavigation of the existing search region. These techniques improve mSHO algorithm's search capabilities, allowing it to navigate the search space and converge toward optimal solutions efficiently. To evaluate the efficacy of the mSHO algorithm, comprehensive assessments are conducted across both the CEC2020 benchmark functions and nine distinct engineering problems. A meticulous comparison is drawn against nine metaheuristic algorithms to validate the achieved outcomes. Statistical tests, including Wilcoxon's rank-sum and Friedman's tests, are aptly applied to discern noteworthy differences among the compared algorithms. Empirical findings consistently underscore the exceptional performance of mSHO across diverse benchmark functions, reinforcing its prowess in solving complex optimization problems. Furthermore, the robustness of mSHO endures even as the dimensions of optimization challenges expand, signifying its unwavering efficacy in navigating complex search spaces. The comprehensive results distinctly establish the supremacy and efficiency of the mSHO method as an exemplary tool for tackling an array of optimization quandaries. The results show that the proposed mSHO algorithm has a total rank of 1 for CEC'2020 test functions. In contrast, the mSHO achieved the best value for the engineering problems, recording a value of 0.012665, 2993.634, 0.01266, 1.724967, 263.8915, 0.032255, 58507.14, 1.339956, and 0.23524 for the pressure vessel design, speed reducer design, tension/compression spring, welded beam design, three-bar truss engineering design, industrial refrigeration system, multi-Product batch plant, cantilever beam problem, multiple disc clutch brake problems, respectively. Source codes of mSHO are publicly available at https://www.mathworks.com/matlabcentral/fileexchange/135882-improved-sea-horse-algorithm

# 1. Introduction

Optimization is the process of reaching the minimum or maximum value of some real function under a limited range of values. Using mathematical notations, the optimization function can be expressed as $f : D \rightarrow R$ from some set $D$ to the real numbers $R$. In a minimization optimization problem, a member $x_0 \in D$, $f(x_0) \leq f(x) \forall x \in A$ whereas in a maximization optimization problem, $f(x_0) \geq f(x) \forall x \in A$ (Pierre, 1986; Smith, 1978). Traditional gradient-based optimization methods rely on finding the derivative of functions, but they have limitations, particularly when dealing with complex optimization problems that lack derivatives or involve many local minima in the search space surface (Sun et al., 2019).

The scientific community has progressively adopted computational intelligence algorithms, such as metaheuristics, to optimize both discrete and continuous problems (S. Chen & Zheng, 2023; Khurma, Aljarah, Sharieh, & Mirjalili, 2020; Xing, Zhao, et al., 2023). Metaheuristic algorithms offer advantages over traditional mathematical algorithms owing to their gradient-free nature, which makes them well-suited for tackling undifferentiated problems and yielding promising near-optimal solutions. Although these solutions may not be optimal, they provide valuable approximations (Hussien et al., 2022; Jia, You, et al., 2023). Furthermore, metaheuristics demonstrate polynomial time complexity, rendering them more efficient than conventional methods with exponential time complexity (Ma et al., 2023; Osaba et al., 2021).

Metaheuristic algorithms have gained prominence in optimizing various problems, primarily due to their derivative-free nature, satisfactory performance metrics, simplicity, efficiency, and robustness (Jia, Shi, et al., 2023; Morales-Castañeda et al., 2020). In the realm of combinatorial optimization, there has been a proliferation of "novel" metaheuristic techniques, many of which draw inspiration from artificial or natural processes. Metaheuristics can be categorized into four main groups, each based on distinct concepts and sources of inspiration: evolutionary algorithms, physics-based algorithms, swarm-based algorithms, and human-based algorithms. These categories encompass a wide range of optimization approaches, each with its unique principles and techniques.

- Evolutionary Algorithms (EAs), like Genetic Algorithms (GA) (Katoch et al., 2021), draw inspiration from biological evolution and natural selection. These algorithms emulate genetic variation, selection, and reproduction processes. GA, for instance, employs a population of potential solutions that evolve over generations using selection, crossover, and mutation operations. This iterative process gradually improves the population's fitness, guiding the optimization procedure by exploring the search space.
- Physics-based algorithms (PhAs) are inspired by fundamental principles and natural phenomena, using simulations of physical processes to optimize solutions. Simulated Annealing (SA) is a well-known example, mimicking the annealing process in metallurgy. SA begins with high "temperature" to encourage exploration and then gradually reduces it to guide optimization towards better solutions. Other PhAs include the weighted mean of vectors (INFO) (Ahmadianfar et al., 2022), rime optimization algorithm (RIME) (Su et al., 2023), Runge-Kutta method (RUN) (Ahmadianfar et al., 2021), and Fick's law algorithm (FLA) (Hashim et al., 2023). These algorithms draw insights from physics to improve optimization strategies.
- Swarm-based algorithms (SAs), inspired by the collective behaviors of natural swarms like bird flocks and ant colonies, prioritize communication, cooperation, and decentralized decision-making among swarm individuals. A notable example is Particle Swarm Optimization (PSO) (Kennedy & Eberhart, 1995), which simulates particle movement and information sharing within a swarm to guide the search process. PSO maintains a balance between exploration and exploitation by adjusting particle velocities based on individual and global best positions, harnessing the collective intelligence of the swarm to explore and exploit the search space for optimal solutions effectively. Other SAs, such as snake optimizers (SO) (Hashim & Hussien, 2022), spotted hyena optimizer (SHO) (Dhiman & Kumar, 2017), slime mould algorithm (SMA) (Li et al., 2020), and colony predation algorithm (CPA) (Tu et al., 2021), similarly draw inspiration from natural swarming behaviors to enhance optimization techniques.
- Human-based algorithms (HAs) belong to the category of metaheuristic algorithms that draw inspiration from human intelligence and problem-solving methods. These algorithms simulate or replicate human decision-making processes and learning mechanisms. A well-known example is Teaching-Learning-Based Optimization (TLBO) (Rao et al., 2011), which models the interaction between a teacher and students to facilitate knowledge transfer and solution improvement. By harnessing human-inspired approaches, these algorithms aim to enhance optimization and discover effective solutions for complex problems. Other HAs include the Mother optimization algorithm (Matoušová et al., 2023) and Human mental (HM) search, (Mousavirad & Ebrahimpour-Komleh, 2017), each seeking to improve optimization using principles inspired by human behavior and cognition.

SAs are mathematical methodologies inspired by the collaborative behaviors observed in natural animal groups. These algorithms translate the survival and foraging behaviors of swarm members into mathematical equations. In response to the No-Free Lunch (NFL) theorem, researchers have developed and refined numerous SAs by drawing inspiration from different aspects of nature or introducing new variants to address their limitations. These variants involve proposing novel operators or techniques that are integrated with the original algorithm. Some enhancement approaches for SAs include incorporating chaotic maps (Khurma, Aljarah, & Sharieh, 2020a), introducing local search (Ahmed et al., 2023), applying opposition-based learning (OBL) (Khurma et al., 2022; Mostafa et al., 2023), utilizing EA selection operators (Khurma et al., 2021), enhancing EA crossover and mutation operators (Alweshah et al., 2022; Awadallah et al., 2022), leveraging Levy flight behavior (Ewees et al., 2022; Mostafa et al., 2022), using Gaussian operators (X. Zhang et al., 2020), and applying rank-based methods (Khurma, Aljarah, & Sharieh, 2020b). These efforts contribute to the ongoing evolution and advancement of SAs for optimization.

---

Various original and enhanced SAs have found applications in diverse fields. For instance, the mCOOT (COOT) algorithm was improved and employed in estimating unmeasured battery parameters, resulting in enhanced accuracy and reduced error rates (Houssein, Hashim, et al., 2022). Similarly, the electrostatically charged particles (ECPO) algorithm was improved and tested on IEEE CEC 2017 test functions, demonstrating its effectiveness in estimating parameters for photovoltaic models (Kamel et al., 2022). In the context of electrical distribution networks, a modified robust optimization method (mFBI) was proposed to optimize the distribution of distributed generators (DGs), leading to reduced energy losses (Tolba et al., 2022). In another study, an improved version of the artificial ecosystem optimization algorithm with opposition-based learning (AEO-OBL) was developed to determine the optimal distribution of DGs in radial distribution networks (Khasanov et al., 2023). This approach considers the stochastic nature of renewable energy sources, like wind turbines and photovoltaic power generation, using appropriate probability models. The loss sensitivity index is utilized to identify suitable buses for integrating DG modules into the network. Additionally, an improved algorithm called Lévy Flight Distribution with opposition-based learning (LFD-OBL) was proposed to address the limitations of the original LFD algorithm. This improved version was applied to optimize the parameters of a Three-Diode Photovoltaic model and demonstrated superior performance (Houssein, Rezk, et al., 2022). These studies highlight the effectiveness and versatility of enhanced SAs in addressing complex optimization problems across different domains.

Metaheuristics have emerged as powerful tools in medical applications, offering innovative solutions in diverse areas. In the context of feature selection, metaheuristic approaches have been harnessed to efficiently identify relevant features from complex medical datasets, aiding in disease diagnosis, prognosis, and treatment planning. These algorithms navigate through high-dimensional data spaces to extract essential information, enhancing the accuracy of predictive models and reducing computational overhead. For example, piri et al. (Piri & Mohapatra, 2021) introduced a novel approach called Multi-Objective Quadratic Binary Harris Hawks Optimization (MOQBHHO), which utilizes the K-Nearest Neighbor (KNN) method as a wrapper classifier. This technique aims to extract optimal feature subsets from medical data for enhanced performance. Thawkar et al. (Thawkar et al., 2021) introduced a hybrid feature selection approach by combining the Butterfly optimization algorithm (BOA) and the Ant Lion optimizer (ALO) to create the hybrid BOAALO method. This method effectively selects an optimal subset of features, which is then employed to predict the benign or malignant status of breast tissue.

Additionally, metaheuristics find utility in multilevel threshold segmentation of medical images, facilitating the precise delineation of anatomical structures or pathological regions. By optimizing threshold values, these algorithms enable accurate segmentation, which is vital for quantitative analysis, disease quantification, and treatment evaluation. The versatility of metaheuristics in handling intricate and often noisy medical data underscores their potential to drive advancements in medical imaging, diagnosis, and patient care. For example, Chakraborty et al. (Chakraborty, Saha, Nama, & Debnath, 2021) focuses on developing a computational tool to quickly and accurately assess illness severity using COVID-19 chest X-ray images. It introduces a modified whale optimization method, named mWOAPR, that enhances diagnostic precision by integrating random population initialization during global search and optimizing parameter settings for improved exploration-exploitation balance. Xing et al. (Xing, Zhao, et al., 2023) introduced an enhanced Whale Optimiza-tion Algorithm (WOA), termed Quasi-Opposition-Based WOA (QGB-WOA), tailored for COVID-19 applications. QGBWOA integrates quasi-opposition-based learning for improved solution search and a Gaussian barebone mechanism to enhance solution space diversity. This refinement holds promise for precise feature selection and multi-threshold image segmentation in COVID-19-related tasks.

In late 2022, a team of researchers introduced the Sea Horse Optimizer (SHO), drawing inspiration from seahorses' locomotion, predation, and reproductive behaviors (Zhao, Zhang, Ma, & Wang, 2022). Seahorses exhibit distinctive locomotion, such as jumping and wrapping their tails around algae or leaves, often influenced by marine eddies, leading to spiral movement. They can also exhibit Brownian motion by turning upside down. Moreover, seahorses employ their uniquely shaped heads to stealthily approach and capture prey, achieving a remarkable success rate of up to 90%. Additionally, the random mating of male and female seahorses contributes to a new generation inheriting advantageous traits from their parents.

The observed seahorse behaviors have endowed the SHO algorithm with the capacity to effectively manage the exploration and exploitation phases when seeking optimal solutions. This ability enables SHO to strike a balance between thorough exploration of the solution space and efficient exploitation of promising areas. Furthermore, the incorporation of these behaviors promotes increased diversity among solutions within the SHO community. Consequently, SHO exhibits enhanced performance by mitigating premature convergence and avoiding getting trapped in local minima. SHO's advantageous attributes have been demonstrated through successful applications in diverse domains. Notably, it has proven effective in tasks such as fine-tuning power system stability and optimizing parameters (Aribowo, 2023). Additionally, SHO has been applied to reduce exhaust pollutants from diesel engines, showcasing its versatility and practical utility (Alahmer et al., 2023).

The unique characteristics and accomplishments of the SHO algorithm have motivated us to extend its application to address a diverse array of engineering challenges. Nevertheless, a notable drawback of SHO lies in its exploitation strategy, which depends on selecting a neighboring individual randomly during local searches within the search space. Recognizing this limitation, we have undertaken the task of enhancing the local search procedure of SHO. To rectify this deficiency, we propose novel methods aimed at optimizing the local search process within the algorithm. These methods are designed to enhance SHO's performance by effectively circumventing the risk of becoming trapped in local minima and facilitating the convergence toward optimal solutions. Our study's primary contributions encompass the following key points:

- Proposing a robust, high-performance variant of SHO, named the mSHO method, enhances the SHO exploitation strategy. This is done by replacing the original method with a new local search strategy, which is done in three steps:

  - Neighborhood-based local search strategy
  - A global non-neighbour-based search strategy
  - Walk around the existing search strategy

- The mSHO method is compared with the original SHO and eight different optimizers in ten CEC2020 test positions.
- mSHO is used to solve nine real-world engineering problems, namely: Welded beam design problem, Three-bar truss design problem, Tension/compression spring design, Speed reducer design, Industrial refrigeration System, Pressure vessel design,

Cantilever Beam Design, Multidisc clutch brake, and Multi-product batch plant.

· Results of mSHO outperformed other algorithms in both constrained & unconstrained problems.

The rest of the paper is structured as follows. Sec. 2 presents some recent literature in which researchers proposed improvements to SAs for solving complicated engineering problems. Sec. 3 describes the SHO algorithm's inspiration and mathematical methodology in detail. Sec. 4 discusses the proposed mSH in detail. Sec. 5 provides the results and in-depth discussion of mSHO and other competitive algorithms on CEC2020 test functions. The mSHO's performance on various engineering problems is presented in Sec. 6. Sec. 7 summarises the results and discusses the limitations of the work. Sec. ?? concludes the paper and offers some potential research directions that can help improve SHO performance as well.

## 2. Related works

This section provides an overview of recent studies that have proposed methods to enhance the performance of specific metaheuristic algorithms for solving global and engineering problems over the past three years. Table 1 provides a summary of the studies mentioned, considering four key criteria: the publication year, the metaheuristic algorithm employed, the enhancement approach taken, whether the IEEE CEC suite was used, the number of benchmark test functions examined, and the quantity and nature of the engineering problems used for evaluation.

Hongwei et al. (Hongwei et al., 2019) utilized chaos theory to introduce an enhanced variant of Moth Flame Optimization (MFO) called CMFO. Chaotic functions were employed for initializing individuals, managing overrides, and adjusting the distance parameter. CMFO underwent testing on three standard function groups and two real-world engineering problems. The statistical findings demonstrated that incorporating an appropriate chaotic map (Singer's map) into the relevant component of MFO significantly improved its performance. Nevertheless, the study did not explore the application of other chaotic maps to the MFO algorithm. Sheikhi et al. introduced an enhanced variant of Temporal Evolutionary Optimization (TEO) called ETEO in their study (Sheikhi Azqandi et al., 2020). This enhancement strategy incorporated a temporal evolution factor and population clustering. A memory was utilized to store some of the best designs. ETEO underwent evaluation across a range of constrained and unconstrained problems, as well as engineering design problems. ETEO aimed to improve TEO's performance, address its weaknesses, and enhance search capabilities during both the exploration and exploitation phases. By employing population clustering, enhancing the environmental factor, and incorporating a memory to preserve some of the best design variables, ETEO demonstrated competitiveness with other metaheuristic algorithms in terms of statistical outcomes, particularly concerning the best objective function and the number of function evaluations performed during optimization.

Chen (H. Chen et al., 2020) introduced OBCWOA, an enhanced variant of the Whale Optimization Algorithm (WOA), which incorporated chaos and quasi-opposition strategies for global optimization problems. OBCWOA demonstrated robustness in solving global optimization problems, excelling in convergence accuracy, speed, high-dimensional search capability, and stability. It was also effective in addressing real engineering design problems. However, OBCWOA had some drawbacks, including the need for adjusting more parameters than the original WOA, longer running times compared to some metaheuristic algorithms, and limited improvement in the correctness of certain test functions. Nonetheless, OBCWOA remained a valuable tool for complex practical problems and remained competitive among state-of-the-art algorithms. In (Fan et al., 2020), ESSAWOA was developed by integrating WOA with SSA and a Lens OBL(LOBL) strategy for global optimization. The exploitative power of the SSA leader's strategy was used to update personnel attitudes before WOA operations were implemented. Subsequently, the nonlinear parameter of SSA in the prey encircling and attacking phases was incorporated into WOA to enhance the convergence behavior. The LOBL strategy was adopted to increase population diversity. ESSAWOA was evaluated using twenty-three standard functions and three classical engineering design problems. The findings show that ESSAWOA can swiftly and efficiently find a promising solution to these optimization issues. ESSAWOA performs much better than the fundamental WOA, SSA, and other meta-heuristic algorithms.

Nadimi et al. (Nadimi-Shahraki et al., 2021) proposed an improved Gray Wolf (IGWO) optimizer for engineering problems. IGWO adopted a new movement strategy called Learning-Based Hunting (DLH) inspired by wolves' natural hunting behavior. DLH implemented a different method of neighborhood identification for each individual so that neighborhood information could be shared between individuals. The performance of the IGWO algorithm was evaluated on a set of CEC 2018 standards and four engineering problems. I-GWO is compared across all tests to six more cutting-edge metaheuristics. Friedman and MAE statistical tests are also used to assess the results. In comparison to the algorithms employed in the studies, the I-GWO algorithm is very competitive and frequently superior, as shown by the experimental findings and statistical testing. The suggested algorithm's performance and applicability on engineering design challenges are shown by the findings.

Wang et al. (Wang et al., 2021) proposed a new variant of the BOA called MBFPA by hybridizing it with the flower pollination and symbiosis mechanism of global optimization problems. Flower pollination and symbiotic organisms support exploration and exploitation capacity, respectively. Moreover, the possibility of alternating exploration and adaptive exploitation improves the balance between these two phases. The MFPPA is tested on 49 standardized test functions and five classic engineering problems. The findings demonstrate the viability of the suggested method and demonstrate its competitiveness and high application prospects. Chakraborty et al.(Chakraborty, Saha, Sharma, et al., 2021) enhanced WOA (WOAmM) was proposed using the mutualism phase from symbiotic organisms search (SOS). The proposed WOAmM method was tested on 36 benchmark functions and IEEE CEC 2019 function suite. In addition, six real-world engineering optimization problems were solved by the proposed method. In comparison to other competing approaches, the results show that the suggested SSC algorithm is resilient, effective, efficient, and convergence analysis.

Zhang (Y. Zhang et al., 2021) improved the global search phase of JAYA and implemented the enhanced Jaya (EJAYA) for global optimization. EJAYA had many distinguished features such as local exploitation, which defined upper and lower local attractors. Furthermore, the global exploration was guided by historical population, and it didn't make any adjustments for initial parameters. The EJAYA was verified by testing it on 45 test functions from IEEE CEC 2014 and IEEE CEC 2015 test suites. Furthermore, EJAYA was implemented to solve seven real-world engineering design optimization problems. The effectiveness of the newly proposed improved techniques to JAYA and the strong ability of EJAYA to escape from the local optimum for tackling difficult optimization issues are supported by experimental results. In (Yıldız et al., 2022),

Yildiz proposed a chaotic RUN (CRUN). In this study, 10 different chaotic maps were integrated into the RUN algorithm to boost its performance, and it was tested on some design engineering problems. The results showed that CRUN was the best compared to the most recent algorithms in the literature. The proposed CRUN method can also uncover advantageous features in a variety of managerial implications, including supply chain management, business models, fuzzy circuits, and management models.

Sharma et al. (Sharma et al., 2022) proposed a new variant of BOA, namely mLBOA. He integrated the self-adaptive parameter setting, Lagrange interpolation formula, a new local search strategy, and levy flight operators with BOA. The IEEE CEC2017 benchmark suite and three real-world engineering design problems were used to evaluate the mLBOA. The outcomes were contrasted with six cutting-edge algorithms and five BOA variations. Additionally, a number of statistical tests have been carried out to support the rank, significance, and complexity of the proposed mLBOA, including the Friedman rank test, Wilcoxon rank test, convergence analysis, and complexity analysis. The mLBOA has also been used to resolve three actual engineering design issues. According to all of the analyses, the suggested mLBOA algorithm is competitive with other well-known state-of-the-art algorithms and BOA variants.

Saha (Saha, 2022) introduced an enhanced version of the Sine Cosine Algorithm (SCA) called MAMSCA. The enhancement involved dividing the SCA's population into two halves and applying either a sine or cosine method to update each half. Furthermore, a modified mutualism phase was incorporated into the algorithm. MAMSCA was applied to standard benchmark functions, IEEE CEC 2019 functions, and five engineering design problems. The results demonstrated significant improvements in addressing real-world problems. A comprehensive assessment of the algorithm, including a statistical analysis, evaluation of time complexity, and solution generation speed, underscored its enhanced performance and suitability for practical applications. Chakraborty (Chakraborty et al., 2023) introduced a novel variant of WOA called m-SDWOA, which integrates WOA with the modified mutualism phase of SOS, the mutation strategy of Differential Evolution (DE), and the commensalism phase of SOS. The algorithm incorporates a new parameter, denoted as $Y$, to determine whether to apply the global or local phases. The efficiency of the algorithm was assessed using 42 benchmark functions, an IEEE CEC 2019 test suite, and four engineering design problems. These evaluations consistently demonstrated the superior performance of the proposed algorithm compared to the methods it was benchmarked against.

In the studies mentioned above, researchers introduced various operators and techniques to address limitations commonly associated with metaheuristic algorithms, including early convergence, bias towards local minima, and imbalances in exploration and exploitation. These innovations have the potential to enhance the optimizer's performance, stability, and robustness, leading to more dependable and effective results. Building upon the foundations of the NFL theory, this paper extends this research trajectory by harnessing the recently developed SHO algorithm. Additionally, the paper integrates specific local search strategies into SHO to further enhance its effectiveness in tackling global optimization and engineering problems.

**Table 1:** Summary of related literature on improved metaheuristics for global and engineering optimization problems

| Ref | Year | Metaheuristc | Improved MA | Improvement | IEEE CEC | #Benchmark functions | #Eng. problems | Name of engineering problems |
|---|---|---|---|---|---|---|---|---|
| Hongwei et al., 2019 | 2019 | MFO | CMFO | Chaotic maps | CEC 2005 | 18 | 2 | Welded beam design, Tensional/compressional spring design |
| Sheikhi Azqandi et al., 2020 | 2020 | TEO | ETEO | Population clustering, Memory-based | Benchmark functions | 54 | 5 | Three-bar truss design, Pressure vessel design, Speed reducer design, Tension/compression spring design, Welded beam design |
| H. Chen et al., 2020 | 2020 | WOA | OBCWOA | Chaos, Quasi-opposition | Benchmark functions | 20 | 1 | Tension/compression string design |
| Fan et al., 2020 | 2020 | WOA | ESSAWOA | SSA, LOBL | Benchmark functions | 23 | 3 | Tension/compression spring design, Pressure vessel design, Welded beam design |
| Nadimi-Shahraki et al., 2021 | 2021 | GWO | IGWO | DLH | CEC 2018 | 30 | 4 | Pressure vessel design, Welded beam design, Optimal power flow( IEEE 118-bus, IEEE 30-bus) |
| Wang et al., 2021 | 2021 | BOA | MBFPA | FPA , Mutation | Benchmark functions | 49 | 5 | Three-bar truss design, Speed reducer, Multi-plate disc clutch brake design, Pressure vessel design, Welded beam design |
| Chakraborty, Saha, Sharma, et al., 2021 | 2021 | WOA | WOAmM | SOS | CEC 2019 | 36 | 6 | Three-bar truss design problem, Gas transmission compressor design, Pressure vessel design, Cantilever beam design, Gear train design |
| Y. Zhang et al., 2021 | 2021 | JAYA | EJAYA | Local exploitation, Global exploration | CEC 2014, CEC 2015 | 45 | 7 | welded beam design, tension/compression spring design, pressure vessel design, speed reducer design, rolling element bearing design, hydrostatic thrust bearing, car side impact design |
| Yıldız et al., 2022 | 2022 | RUN | CRUN | chaotic maps | | | 6 | gear train design, coupling with a bolted rim, pressure vessel design, Belleville spring, vehicle brake-pedal |

*Continue on the next page*

**Table 1:** Summary of related literature on improved metaheuristics for global and engineering optimization problems

| Ref | Year | Metaheuristc | Improved MA | Improvement | IEEE CEC | #Benchmark functions | #Eng. problems | Name of engineering problems |
|---|---|---|---|---|---|---|---|---|
| Sharma et al., 2022 | 2022 | BOA | mLBOA | Self-adaptive parameter setting, Lagrange interpolation, Local search,Levy flight | CEC 2017 | 15 | 3 | Spread Spectrum Radar Polly Phase Design, Three-bar Truss Design, Gas Transmission Compressor Design |
| Saha, 2022 | 2022 | SCA | MAMSCA | Dividing the population, Modified mutualism | CEC 2019 | 50 | 5 | Gear train design, Gas transmission compressor design, Car side impact design, Cantilever beam design, Three-bar truss design |
| Chakraborty et al., 2023 | 2023 | WOA | m-SDWOA | SOS, DE | CEC 2019 | 42 | 4 | Gear train design, Gas transmission compressor design, Welded beam design, Weight minimization of a speed reduce |

# 3. Background

## 3.1 Sea Horse Optimizer (SHO)

Sea Horse Optimizer (SHO) draws inspiration from the predation, movement, and breeding behaviors of seahorses, which enable them to adapt to their environment and survive. Seahorses, small fish found in warm waters, have a head resembling that of a horse. In terms of movement, seahorses exhibit a spiral motion when wrapping their tails around a stem (or leaf) of algae. Their unique head shape aids in stealthy predation. Furthermore, seahorses engage in random mating between females and males to produce offspring in their breeding behavior. The algorithm encompasses four phases: initialization, movement behavior, predation behavior, and breeding behavior.

In the initialization phase, the algorithm generates the initial population of sea horses, as represented by Equation 1, where $Dim$ represents the dimension and $pop$ is the population size (Zhao et al., 2023).

$$Seahorses = \begin{bmatrix} x_1^1 & ... & x_1^{Dim} \\ ... & ... & ... \\ x_{pop}^1 & ... & x_{pop}^{Dim} \end{bmatrix} \quad (1)$$

Each individual is represented by Equation 2, with each value in the list calculated using Equation 3, where rand is a random value in the range of [0, 1]. Here, $x_i^j$ represents the $j^{th}$ dimension of the $i^{th}$ individual, and $LB^j$ and $UB^j$ denote the lower and upper bounds of the $j^{th}$ dimension.

$$X_i = [x_i^1, x_i^2, ..., x_i^{Dim}] \quad (2)$$

$$x_i^j = rand \times (UB^j - LB^j) + LB^j \quad (3)$$

The individual with the lowest fitness function value is referred to as the elite individual $X_{elite}$, which is calculated using Equation 4 (Zhao et al., 2023).

$$X_{elite} = argmin(fitness(X_i)) \quad (4)$$

In the movement behavior phase, Sea Horses exhibit two types of movements: the spiral motion and the Brownian motion. When engaged in a spiral motion, the new position of a Sea Horse is determined using Equation 5, where the values of $x$, $y$, and $z$ are computed as shown in Equations 6, 7, and 8. Here, $\rho = u \times e^{\theta v}$ represents the length of the stems defined by the logarithmic spiral constants $u$ and $v$, which are set to 0.05. $\theta$ is a random value within the range [0, 2π]. The Levy distribution function, $Levy(\lambda)$, is calculated using Equation 9, where $\lambda$ is a random number in the range [0, 2], and $s$ is fixed at 0.01. The variables $w$ and $k$ are random numbers in the range [0, 1], and $\sigma$ is determined by Equation 10 (Zhao et al., 2023).

$$X_{new}^1(t+1) = X_i(t) + Levy(\lambda)((X_{elite}(t) - X_i(t)) \times x \times y \times z + X_{elite}(t)) \quad (5)$$

$$x = \rho \times cos(\theta) \quad (6)$$

$$y = \rho \times sin(\theta) \quad (7)$$

$$z = \rho \times \theta \quad (8)$$

$$Levy(\lambda) = s \times \frac{w \times \sigma}{|k|^{\frac{1}{\lambda}}} \quad (9)$$

$$\sigma = \left( \frac{\Gamma(1+\lambda) \times sin(\frac{\pi\lambda}{2})}{\Gamma(\frac{1+\lambda}{2}) \times \lambda \times 2^{\frac{\lambda-1}{2}}} \right) \quad (10)$$

Conversely, in the case of Brownian motion, the new position of a Sea Horse is determined using Equation 11, where $l$ is a constant coefficient. The value of $\beta_t$ is calculated according to Equation 12 (Zhao et al., 2023).

$$X_{new}^1(t+1) = X_i(t) + rand * l * \beta_t * (X_i(t) - \beta_t * X_{elite}) \quad (11)$$

$$\beta_t = \frac{1}{\sqrt{2n}} exp(-\frac{x^2}{2}) \quad (12)$$

To summarize the calculations, Equation 13 encompasses the calculations of the new positions, with $r_1$ denoting a random number (Zhao et al., 2023).

$$X_{new}^1(t+1) = \begin{cases} X_i(t) + Levy(\lambda)((X_{elite}(t) - X_i(t)) \times x \times y \times z + X_{elite}(t)) & r_1 > 0 \\ X_i(t) + rand * l * \beta_t * (X_i(t) - \beta_t * X_{elite}) & r_1 \le 0 \end{cases} \quad (13)$$

The predation behavior is calculated using Equation 14, where $\alpha$ is determined as shown in Equation 15, and $r_2$ represents a random number within the range [0, 1] (Zhao et al., 2023).

$$X_{new}^2(t+1) = \begin{cases} \alpha * (X_{elite} - rand * X_{new}^1(t)) + (1 - \alpha) * X_{elite} & r_2 > 0.1 \\ (1 - \alpha) * (X_{new}^1(t) - rand * X_{elite}) + \alpha * X_{new}^1(t) & r_2 \le 0.1 \end{cases} \quad (14)$$

$$\alpha = (1 - \frac{t}{T})^{\frac{2t}{T}} \quad (15)$$

The breeding behavior is determined by assigning roles to mother and father sea horses, as depicted in Equations 16 and 17, where $X_{sort}^2$ signifies all $X_{new}^2$ sorted in ascending order of their fitness values (Zhao et al., 2023). The actual mating process to produce new offspring is described in Equation 18, where $r_3$ is a random number within the range [0, 1], $i$ is a positive integer within the range [1, $pop$/2], and $X_i^{father}$ and $X_i^{mother}$ represent randomly selected father and mother individuals (Zhao et al., 2023).

$$fathers = X_{sort}^2(1 : pop/2) \quad (16)$$

$$mothers = X_{sort}^2(pop/2 + 1 : pop) \quad (17)$$

$$X_i^{offspring} = r_3 X_i^{father} + (1 - r_3) X_i^{mother} \quad (18)$$

# 4. Proposed method

The original SHO algorithm exhibits certain shortcomings, particularly in achieving a harmonious balance between global and local search behaviors during the movement phase. This issue arises from the random selection of the search strategy, whether it's Spiral or Brownian motion, based solely on a random num-ber $r_1$. Furthermore, the fixed values assigned to parameters $u$ and $v$, which dictate the length of the stems, remain constants throughout the optimization process, potentially impeding the algorithm's ability to guide solutions effectively to new positions. To address these limitations, this paper introduces an improved version of SHO, named mSHO, aimed at enhancing the algorithm's performance and addressing its main limitations.

In this section, we delve into the proposed mSHO method, which brings about significant changes in the movement behavior phase. Instead of the traditional approach, the mSHO method incorporates the following three distinct steps:

· Neighborhood-based local search strategy.
· Non-neighborhood based global search strategy.
· Wandering around based search strategy.

Neighborhood-Based Local Search Strategy leverages an individual's conscious neighborhood to enhance the quality of exploitation within that neighborhood. Specifically, a random neighbor, denoted as $c_{local}$, is chosen from within the individual's local neighborhood, and another neighbor, termed $c_{global}$, is selected from outside the local neighborhood but possesses the lowest fitness function value. Subsequently, if the fitness value of $c_{local}$ is found to be lower than that of $c_{global}$, the individual adjusts its position towards that of $c_{local}$, as calculated by Equation 19 (Zamani et al., 2019).

$$X_i(t+1) = X_i(t) + r_i \times fl_i(t) \times (m_{local}(t) - X_i(t)) \quad (19)$$

where $fl_i(t)$ is the flight length of the individual in iteration $t$, $r_i$ is a random number in the range of [0, 1], and $m_{local}(t)$ is the hiding position of $c_{local}$ for iteration $t$.

In contrast, the Non-neighborhood based Global Search Strategy is activated when the fitness value of $c_{local}$ exceeds or equals the fitness value of $c_{global}$. In this situation, the individual moves toward the position of $c_{global}$, represented as $X_{ij}(t+1)$, and this relocation is determined using Equation 20 (Zamani et al., 2019).

$$X_{ij}(t+1) = r_i \times fl_i(t) \times (m_{globalj}(t) - X_{ij}(t)) \quad (20)$$

where $j$ is the dimension value, $m_{globalj}(t)$ is the hiding position of $c_{global}$ for iteration $t$ and dimension $j$.

The Neighborhood-based Local Search Strategy and the Non-neighborhood based Global Search Strategy both include a validation step to ensure that the new position falls within the problem space's defined range. If it does not, the strategy randomly adjusts the dimensions that have exceeded this range, bringing them back into the problem space's boundaries.

On the other hand, the Wandering Around-Based Search Strategy is employed when the previous two strategies fail to improve an individual's fitness value. It operates by analyzing the surrounding environment and maneuvering the individual to a potentially more favorable position with a lower fitness value. Equation 21 calculates this new position, where $m_{gbestj}(t)$ represents the best hiding position for dimension $j$, and $X_{rj}(t)$ corresponds to a randomly selected individual in the $j^{th}$ dimension (Zamani et al., 2019).

$$X_{ij}(t+1) = m_{gbestj}(t) + r_i \times fl_i(t) \times (X_{rj}(t) - X_{ij}(t)) \quad (21)$$

Figure 1 illustrates the step-by-step process of the proposed mSHO algorithm. The algorithm commences by generating the initial population of Sea Horses, following the principles outlined in Equations 1, 2, and 3. In each iteration, the algorithm proceeds to evaluate the fitness of each individual and updates the elite individual using Equation 4. Subsequently, for each individual, two neighboring Sea horses, denoted as $c_{local}$ and $c_{global}$, are selected. Their fitness values are then compared, and the individual adopts the position of the Sea Horse with the lower fitness value. This position update is determined by Equations 19 and 20. Following this, another fitness comparison is conducted, this time between the individual's fitness at the new position and its fitness at the previous position. If the fitness at the new position is not lower than the previous one, the individual's position is modified using Equation 21. The predation and breeding behaviors are calculated according to Equations 14. This flowchart provides a comprehensive overview of the mSHO algorithm's operation.

# 5. Assessment of mSHO on CEC2020 test functions

To prove the efficiency of mSHO, several tests and experiments have been conducted. This study covers two major tests: global optimization problems using ten functions from CEC2020 and nine engineering problems. All experiments were run using MAT-LAB 2022b on an Intel Core$^{TM}$ i7 (3.40 GHz) CPU with RAM 16GB running Microsoft Windows 11.

Several metaheuristics were evaluated and compared with the proposed mSHO in this experiment to ensure a fair assessment. The selected metaheuristics include Dandelion Optimizer (DO) (Zhao, Zhang, Ma, & Chen, 2022), Covariance Matrix Adaptation Evolution Strategy (CMA-ES) (Hansen & Ostermeier, 2001), Hunger games search (HGS) (Yang et al., 2021), Smell Agent Optimization (SAO) (Salawudeen et al., 2021), Harris Hawks Optimization (HHO) (Heidari et al., 2019), Particle Swarm Optimization (PSO)(Kennedy & Eberhart, 1995), and Stochastic Paint Optimizer (SPO) (Kaveh et al., 2020). All algorithms were evaluated under the same conditions with 30 search agents and a maximum of 1000 iterations. To eliminate the impact of random initialization, 30 independent runs were performed, and the algorithms' performance was evaluated using the average fitness and standard deviation metrics. The parameters of the other algorithms are mentioned in Table 2.

## 5.1 Experimental series 1: CEC2020

In this section, we analyze the outcomes of our experiments on the CEC2020 functions, categorizing our findings into four distinct segments: statistical analysis, boxplot representation, convergence assessment, and the Wilcoxon rank test. The CEC2020 dataset encompasses ten distinct functions, as outlined in Table 3. These functions are classified into four categories: Uni-modal, Multi-modal shifted and rotated functions, hybrid, and composition functions. Each function, denoted as $Fi$, is associated with an optimal value that serves as our objective. For instance, F1's optimal value is set at 100, with the optimizer striving to identify a solution that closely approximates this value.

### 5.1.1 Statistical analysis on CEC'2020 test suite

The fitness function values for the different CEC2020 functions with different competitive algorithms are displayed in Table 4. Each function's mean, standard deviation, and rank are calculated across the different optimization algorithms. The mean value is useful because it provides a comparative estimate of the total of many runs. Standard deviation, on the other hand, provides variability of the different runs. The rank is given based on the mean value, where the rank of 1 is provided to the lowest value.

It is observed from the Table that the proposed mSHO has very competitive fitness values with the rank of 1 for all of the functions except the F6 and F10. The HGS and CMA-ES algorithms have better values for F6 and F10, respectively. In addition, the HGS algorithm has an overall competitive rank of 2 for many functions, while the proposed mSHO algorithm has a total rank of 1. On the other hand, the Friedman test shows that the proposed mSHO displays the lowest value of 1.3, followed by HGS and PSO, whereas SAO has the worst value.

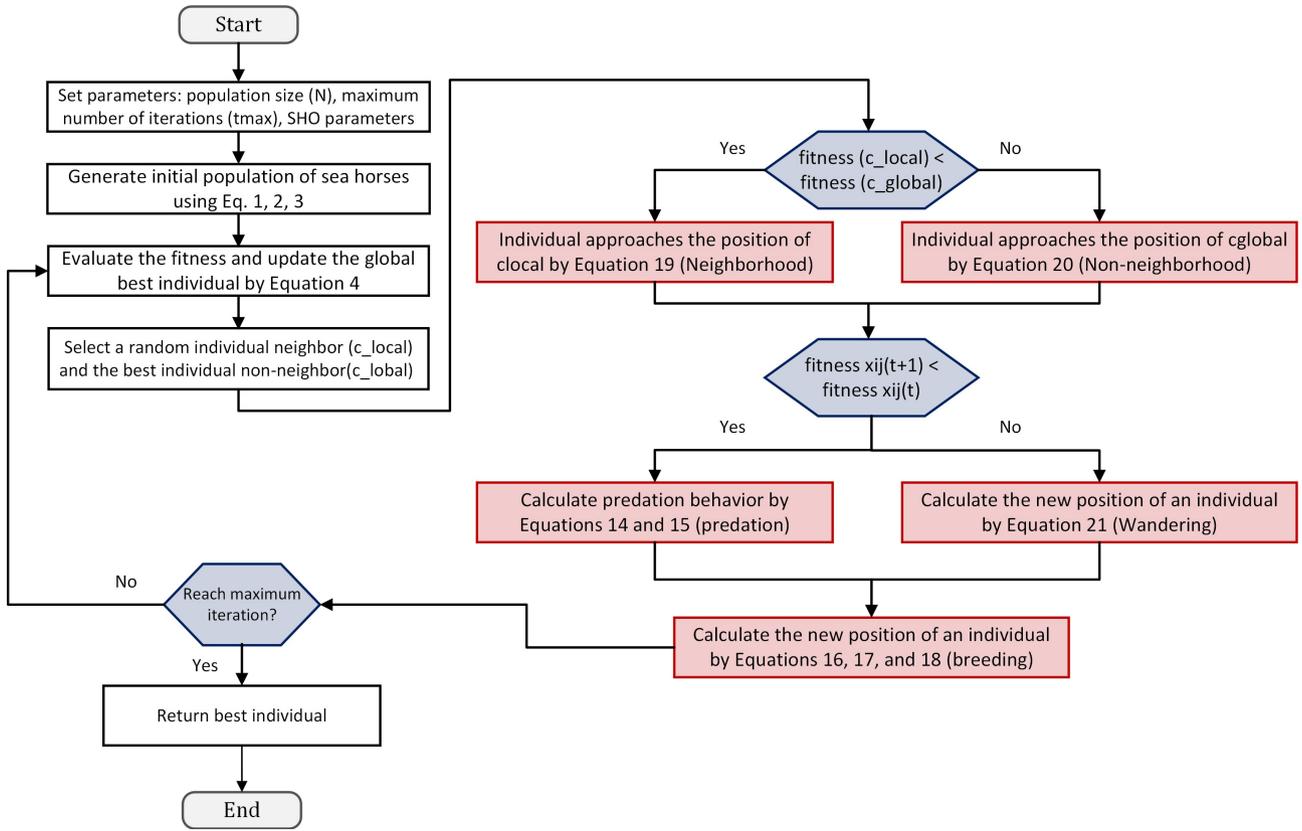

**Figure 1:** Flowchart of the proposed mSHO

**Table 2:** Parameter settings

|  | Parameter | Value |
|---|---|---|
| Population size (N) | CEC2020 problem | 30 |
|  | Engineering problem | 30 |
| Maximum iterations | CEC2020 | 1000 |
|  | Engineering problem | 1000 |
| Problem dimensions (D) | CEC2020 problem | 10 |
|  | Engineering problem | Dimension of problem |
| PSO | Cognitive component (c1) | 2 |
|  | Social component (c2) | 2 |
|  | Inertia weight | 0.2 - 0.9 |
| DO | Adaptive parameters ($\alpha$, $k$) | [0,1], [0,1] |
| HGS | $k$ | 0.3 |
|  | $r_1, r_2, r_3, r_4, r_5, r_6$ | $rand[0, 1]$ |
| SAO | $olf$ | 0.75 |
|  | $SL$ | 0.9 |
| HHO | $\beta$ | 1.5 |
| AOA | $\mu$ | 0.499 |
|  | $\alpha$ | 5 |

**Table 3:** CEC'2020 test suite description

| No. | Function specification | Fi* |
|---|---|---|
| | Uni-modal function | |
| F1 | Shifted and Rotated Bent Cigar Function | 100 |
| | Multi-modal shifted and rotated functions | |
| F2 | Shifted and Rotated Schwefel's Function | 1100 |
| F3 | Shifted and Rotated Lunacek bi-Rastrigin Function | 700 |
| F4 | Expanded Rosenbrock's plus Griewangk's Function | 1900 |
| | Hybrid functions | |
| F5 | N = 3 | 1700 |
| F6 | N = 4 | 1600 |
| F7 | N = 5 | 2100 |
| | Composition functions | |
| F8 | N = 3 | 2200 |
| F9 | N = 4 | 2400 |
| F10 | N = 5 | 2500 |

### 5.1.2  Boxplot behavior analysis

The boxplots of the 30 runs over the different algorithms are presented in Figure 2. The boxplot shows the maximum, minimum, median, and interquartile range (Qaddoura et al., 2021). The proposed mSHO shows compacted box distribution compared to the other algorithms, indicating a stable algorithm.

The proposed mSHO also has the lowest minimum and maximum values for all the functions except for F6. Overall, the proposed mSHO confirms the consistency of the algorithm. Some other observations can be concluded from the figure. SHO has a high standard deviation compared to the other algorithms for F1, while SAO has a high standard deviation for F10 as well as DO for F6 and SOA for F5. F8 shows an interesting pattern since all the algorithms show large values for the standard deviation except for the mso, which indicates a very stable algorithm.

### 5.1.3  Convergence performance analysis

The convergence curve represents the values of the fitness function across the different iterations. This is presented in Figure 3 for the experimented results. The convergence curve is important since it shows that the value of the fitness function decreases when progressing through the iterations. It also shows that after some iterations, the value of the fitness function stays as is, indicating that the algorithm cannot explore better solutions and that the best solution resulting from the algorithm is the closest solution to the correct one.

The convergence curves show advanced values for most functions across all the iterations except for F1 and F6, while F1 obtained the lowest fitness value at the final iterations. This proves the effectiveness of the optimization task for the proposed mSHO algorithm by converging toward minimum values. On the other hand, F2, F5, F7, and F8 show that mSHO is finding better solutions than the other algorithms, but it has similar behavior to the other algorithms for F3, F6, and F10.

### 5.1.4  Wilcoxon rank test analysis

The p-values of the Wilcoxon rank sum test for each competitive algorithm with the proposed mSHO are represented in Table 5. Wilcoxon rank sum test is a non-parametric test to find the significance of the results. It is proposed by (Wilcoxon, 1992) with a 5% significant level. It is observed from the table that the proposed mSHO wins in all comparisons except when compared with DO for F6, HGS for F10, PSO for F8/F10, and SPO for F5.

## 6. Performance of mSHO on engineering design problems

This section evaluates the mSHO algorithm's performance in real-world engineering applications such as:

· Pressure vessel design problem
· Speed beam design problem
· Tension/compression spring design
· Welded beam design problem
· Three-bar truss engineering design problem
· Industrial refrigeration system problem
· Multi-Product batch plant problem
· Cantilever beam problem
· Multidisc clutch brake problem

Those problems have been addressed using mSHO and comparing results against those of other competing algorithms. The mSHO and competing algorithms were run 30 separate times with a total of 1000 iterations to ensure a fair comparison.

### 6.1  Pressure vessel design problem

One of the most common engineering design problems is the Pressure Vessel Design problem, with the aim of finding the cost of the pressure vessel. This problem has four different types of variables: head thickness ($T_h$), shell thickness ($T_s$), length of cylindrical unit ($L$), and the inner radius ($R$). The mathematical structure of the pressure vessel design problem and the four types of constraints applied to the problem design is presented in Equation 22. This engineering problem (tensile design/compressed spring) is solved using the proposed mSHO and other competitive algorithms as shown in Table 6. The obtained statistical results are presented in Table 7. Table 7 shows that the optimal value of the function is 0.012665, which was achieved using the mSHO algorithm. Results reveal that FLA is superior to all other competing algorithms.

In addition, as shown in Figure 4, the convergence curves and boxplot for the mSHO algorithm and other compared methods for the Pressure Vessel Design problem are presented. The convergence curve plots the average best values against the number of iterations for the mSHO algorithm and other compared algorithms after running 1000 times. The results indicate that the mSHO algorithm converges faster than the other algorithms and can typically reach a near-optimal solution more quickly. While the other algorithms demonstrate competitive performance, the SAO and CMA-ES exhibit the lowest performance. Furthermore, the boxplot results illustrate the stability of the proposed mSHO algorithm, followed by the AOA and PSO algorithms. These findings demonstrate the effectiveness and stability of the proposed

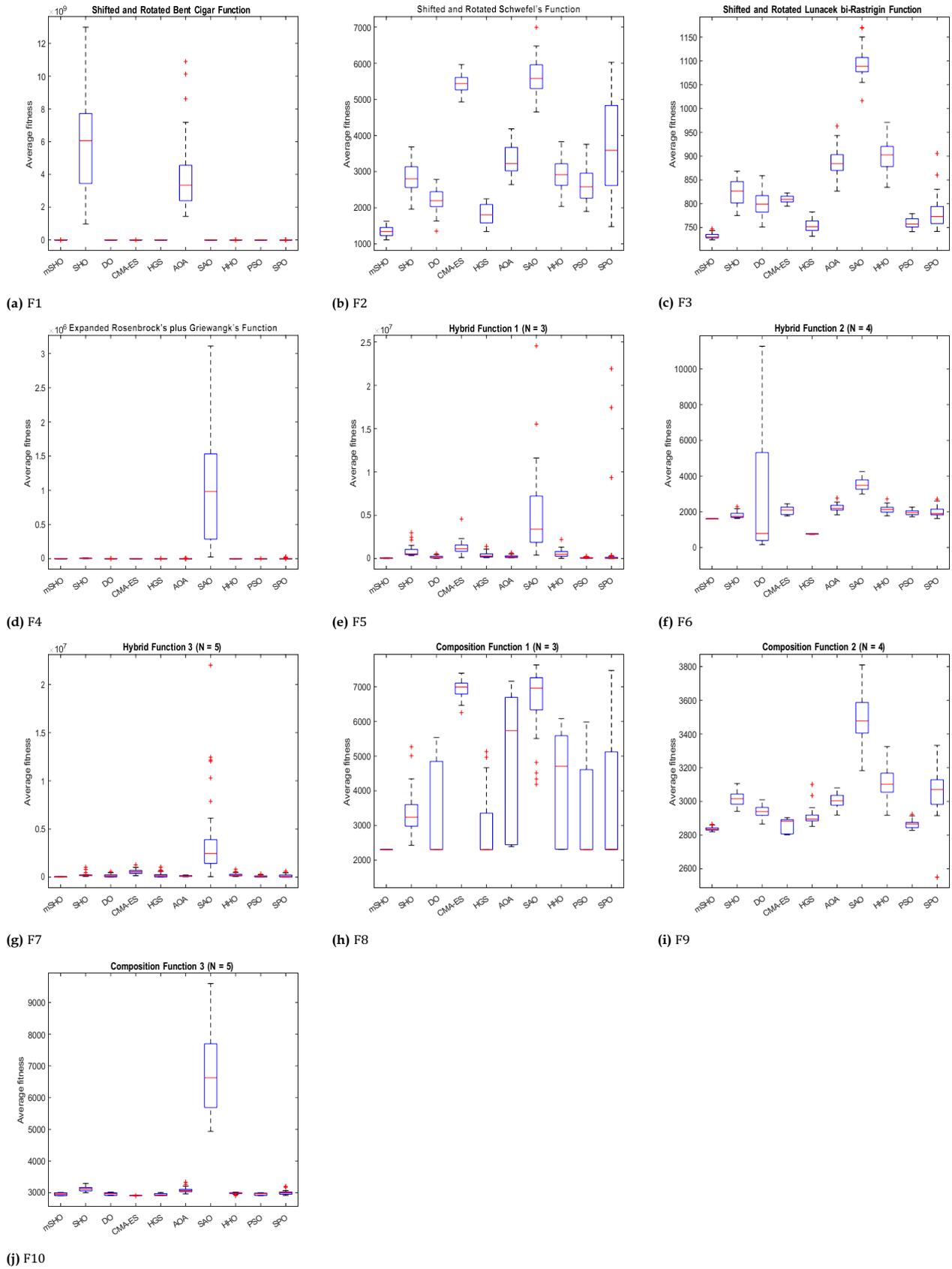

**(a)** F1

**(b)** F2

**(c)** F3

**(d)** F4

**(e)** F5

**(f)** F6

**(g)** F7

**(h)** F8

**(i)** F9

**(j)** F10

**Figure 2:** The boxplot curves of the proposed mSHO and the other approaches obtained over CEC'2020 test suite with $Dim = 10$.

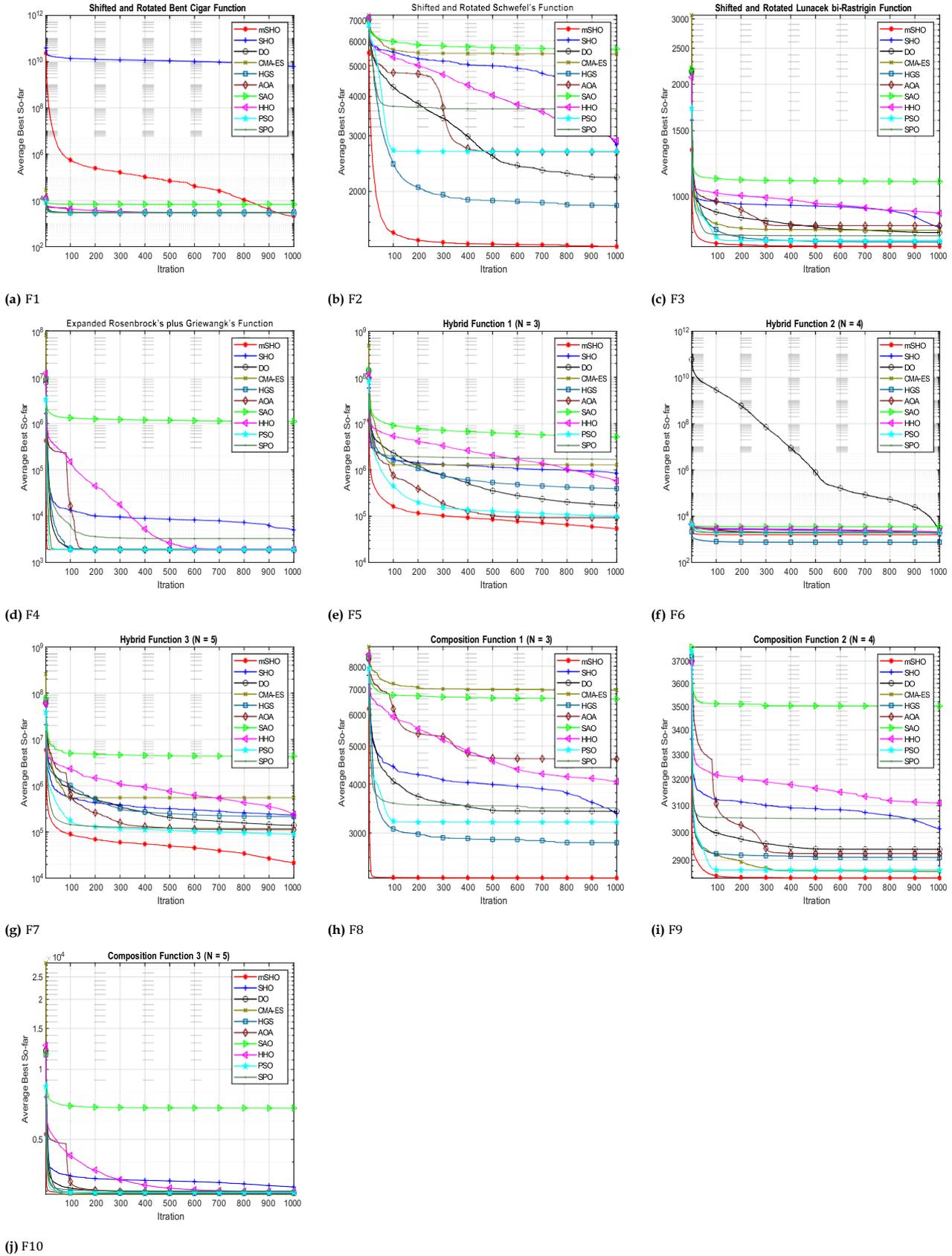

**Figure 3:** The convergence curves of the proposed mSHO and the competitor algorithms obtained on CEC'2020 test suite with *Dim* = 10.

**Table 4:** The mean and STD of fitness values for 30 runs obtained by the competitor algorithms on the CEC'2020 test suite with Dim = 10

| Function | Measures | mSHO | SHO | DO | CMA-ES | HGS | AOA | SAO | HHO | PSO | SPO |
|---|---|---|---|---|---|---|---|---|---|---|---|
| F1 | Mean | **2167.666** | 6.09E+09 | 2960.56 | 2906.272 | 2933.422 | 4.09E+09 | 6813.764 | 2982.666 | 2958.932 | 3001.223 |
| | Std | 2818.008 | 3.05E+09 | 39.30731 | **0.010848** | 32.80232 | 2.5E+09 | 1204.266 | 25.08729 | 31.72322 | 64.70297 |
| | Rank | 1 | 10 | 5 | 2 | 3 | 6 | 9 | 7 | 4 | 8 |
| F2 | Mean | **1339.177** | 2817.837 | 2217.5 | 5457.625 | 1806.443 | 3325.647 | 5650.254 | 2898.974 | 2677.729 | 3653.926 |
| | Std | **149.4172** | 443.9491 | 325.2414 | 279.64 | 281.3332 | 406.6865 | 518.6831 | 463.5156 | 464.5185 | 1202.01 |
| | Rank | 1 | 6 | 3 | 9 | 2 | 4 | 10 | 7 | 5 | 8 |
| F3 | Mean | **731.9907** | 824.6887 | 798.2153 | 810.0422 | 753.2998 | 887.7204 | 1095.175 | 899.9842 | 758.5414 | 782.7039 |
| | Std | **5.659652** | 28.792 | 26.43054 | 7.270733 | 13.70135 | 30.08877 | 33.57846 | 30.11809 | 11.17006 | 35.64884 |
| | Rank | 1 | 7 | 5 | 6 | 2 | 8 | 10 | 9 | 3 | 4 |
| F4 | Mean | **1901.451** | 5076.296 | 1908.204 | 1906.697 | 1903.654 | 2234.568 | 1111983 | 1923.178 | 1902.609 | 3297.193 |
| | Std | **0.373742** | 1953.856 | 4.256352 | 1.816287 | 1.436275 | 772.5102 | 930956 | 5.257264 | 0.885894 | 4944.59 |
| | Rank | 1 | 9 | 6 | 5 | 3 | 4 | 10 | 7 | 2 | 8 |
| F5 | Mean | **54240.62** | 848010.9 | 172393.5 | 1292352 | 397062.9 | 253090.3 | 5219985 | 581330.9 | 99603.21 | 1696510 |
| | Std | **29805.33** | 656376.1 | 136582.4 | 822183.4 | 307375.9 | 152575.9 | 5142683 | 447915.4 | 60581.07 | 5204490 |
| | Rank | 1 | 7 | 4 | 8 | 5 | 2 | 10 | 6 | 3 | 9 |
| F6 | Mean | 1601.524 | 1814.781 | 3080.036 | 2079.857 | **753.2998** | 2222.706 | 3517.05 | 2161.23 | 1931.721 | 1985.644 |
| | Std | **0.416483** | 177.8612 | 3665.986 | 220.6828 | 13.70135 | 201.3467 | 335.3659 | 211.6085 | 131.3297 | 256.4341 |
| | Rank | 2 | 3 | 9 | 7 | 1 | 4 | 10 | 8 | 5 | 6 |
| F7 | Mean | **21683.19** | 233325.4 | 139959.4 | 551833.7 | 215977 | 109998.4 | 4322030 | 242057.3 | 89234.49 | 118786 |
| | Std | **10175.81** | 205427.5 | 143246.4 | 260253.9 | 259901.8 | 42789.46 | 4895366 | 174732.6 | 79457.84 | 151227.1 |
| | Rank | 1 | 7 | 5 | 9 | 6 | 3 | 10 | 8 | 2 | 4 |
| F8 | Mean | **2300.408** | 3364.95 | 3408.826 | 6943.684 | 2834.071 | 4651.941 | 6598.57 | 4060.821 | 3197.561 | 3478.461 |
| | Std | **0.588443** | 642.2368 | 1395.141 | 290.1084 | 946.9044 | 2117.116 | 976.2632 | 1592.76 | 1327.532 | 1708.349 |
| | Rank | 1 | 4 | 5 | 10 | 2 | 8 | 9 | 7 | 3 | 6 |
| F9 | Mean | **2836.961** | 3012.995 | 2938.716 | 2859.627 | 2909.554 | 3006.273 | 3501.35 | 3109.96 | 2864.636 | 3050.984 |
| | Std | **11.55151** | 40.23653 | 36.40562 | 40.10379 | 51.90887 | 42.96402 | 139.0932 | 92.04638 | 25.1554 | 135.1523 |
| | Rank | 1 | 7 | 5 | 3 | 4 | 6 | 10 | 9 | 2 | 8 |
| F10 | Mean | 2948.073 | 3122.153 | 2960.56 | **2906.272** | 2933.422 | 3081.367 | 6813.764 | 2982.666 | 2958.932 | 3001.223 |
| | Std | 37.95805 | 83.68379 | 39.30731 | **0.010848** | 32.80232 | 76.68404 | 1204.266 | 25.08729 | 31.72322 | 64.70297 |
| | Rank | 3 | 9 | 5 | 1 | 2 | 6 | 10 | 7 | 4 | 8 |
| Friedman's mean rank | | 1.3 | 6.8 | 5 | 5.7 | 3 | 6.8 | 9.7 | 7 | 3.1 | 6.6 |
| Rank | | 1 | 7 | 4 | 5 | 2 | 7 | 9 | 8 | 3 | 6 |

mSHO algorithm in tackling the pressure vessel design problem.

Consider $\vec{x} = \begin{bmatrix} x_1 & x_2 & x_3 & x_4 \end{bmatrix} = \begin{bmatrix} T_s & T_h & R & L \end{bmatrix}$,

Minimize $0.6224 x_1 x_3 x_4 + 1.7781 x_2 x_3^2 + 3.1661 x_1^2 x_4 + 19.84 x_1^2 x_3$,

Subject to $g_1(\vec{x}) = -x_1 + 0.0193 x_3 0$,

$g_2(\vec{x}) = -x_2 + 0.00954 x_3 0$,

$g_3(\vec{x}) = -\Pi x_3^2 x_4 - \frac{4}{3} \Pi x_3^3 + 12960000$,

$g_4(\vec{x}) = x_4 - 2400$

Variables range $0 x_1 99$,

$0 x_2 99$,

$10 x_3 200$

$10 x_4 200$

(22)

## 6.2 Speed reducer design problem

One of the most significant engineering design problems is the speed reducer, as described in the study by Sadollah et al. (Sadollah et al., 2013). The primary goal of this problem is to minimize the weight of the speed reducer by optimizing seven variables while also accounting for limitations on the curvature stress of gear teeth, transverse deflections of the shafts, stresses in the shafts, and surface stress. The mathematical model for this problem is presented below:

**Table 5:** mSHO vs. other meta-heuristics algorithms for CEC2020 (D =10) in terms of p-values of the Wilcoxon ranksum test

| mSHO vs. | SHO | DO | CMA-ES | HGS | AOA | SAO | HHO | PSO | SPO |
|----------|-----|-----|--------|-----|-----|-----|-----|-----|-----|
| F1 | 3.02E-11 | 0.007959 | 0.0079312 | 0.007958996 | 3.01986E-11 | 5.53286E-08 | 0.007959 | 0.007959 | 0.007959 |
| F2 | 3.02E-11 | 1.09E-10 | 3.02E-11 | 1.3111E-08 | 3.01986E-11 | 3.01986E-11 | 3.02E-11 | 3.02E-11 | 4.5E-11 |
| F3 | 3.02E-11 | 3.02E-11 | 3.02E-11 | 4.57257E-09 | 3.01986E-11 | 3.01986E-11 | 3.02E-11 | 9.92E-11 | 4.5E-11 |
| F4 | 3.02E-11 | 3.02E-11 | 4.077E-11 | 3.15889E-10 | 3.01986E-11 | 3.01986E-11 | 3.02E-11 | 3.65E-08 | 3.02E-11 |
| F5 | 3.02E-11 | 1.53E-05 | 3.02E-11 | 4.50432E-11 | 1.32885E-10 | 3.01986E-11 | 3.5E-09 | 0.00073 | 0.982307 |
| F6 | 3.02E-11 | 0.379036 | 3.02E-11 | 3.01986E-11 | 3.01986E-11 | 3.01986E-11 | 3.02E-11 | 3.02E-11 | 3.02E-11 |
| F7 | 3.02E-11 | 1.61E-06 | 3.02E-11 | 7.59915E-07 | 3.01986E-11 | 3.01986E-11 | 3.02E-11 | 0.000132 | 0.002755 |
| F8 | 3.02E-11 | 9.26E-09 | 3.02E-11 | 0.000117472 | 3.01986E-11 | 3.01986E-11 | 3.02E-11 | 0.137241 | 6.07E-11 |
| F9 | 3.02E-11 | 3.02E-11 | 0.0292054 | 6.69552E-11 | 3.01986E-11 | 3.01986E-11 | 3.02E-11 | 2.32E-06 | 5.57E-10 |
| F10 | 3.69E-11 | 0.030317 | 6.715E-05 | 0.673495053 | 1.95678E-10 | 3.01986E-11 | 0.000318 | 0.122353 | 0.000125 |

**Table 6:** Best solution obtained from the comparative algorithms for solving the pressure vessel design problem

| Algorithm | x1 | x2 | x3 | x4 | Cost |
|-----------|-----|-----|-----|-----|------|
| mSHO | 0.774555 | 0.383203 | 40.31962 | 200 | 5870.12409 |
| SHO | 0.783011 | 0.395825 | 40.70835 | 194.6584 | 5908.30062 |
| DO | 0.774533 | 0.383229 | 40.31962 | 200 | 5870.12562 |
| CMA-ES | 0.859051 | 0.473962 | 43.62445 | 180.7057 | 6879.70268 |
| HGS | 0.774549 | 0.383204 | 40.31962 | 200 | 5870.12398 |
| AOA | 0.774549 | 0.383204 | 40.31962 | 200 | 5870.12398 |
| SAO | 2.527902 | 5.801292 | 1.559886 | 5.133647 | 7710.96822 |
| HHO | 0.835827 | 0.430123 | 43.61014 | 158.7634 | 6046.19065 |
| PSO | 0.778476 | 0.385079 | 40.52191 | 197.203 | 5876.94102 |
| SPO | 0.774574 | 0.383204 | 40.31962 | 200 | 5870.1246 |

**Table 7:** Results obtained from competitor algorithms for pressure vessel design problem

| Mea. | mSHO | SHO | DO | CMA-ES | HGS | AOA | SAO | HHO | PSO | SPO |
|------|------|-----|-----|--------|-----|-----|-----|-----|-----|-----|
| Min | 5870.1241 | 5908.3006 | 5870.1256 | 6879.703 | 5870.124 | 5870.124 | 7710.968 | 6046.191 | 5876.941 | 5870.125 |
| Max | 5870.1342 | 7271.3656 | 7156.128 | 8758.232 | 7301.196 | 6666.339 | 23390.18 | 7464.369 | 6898.222 | 21829.09 |
| Mean | 5870.1266 | 6196.8232 | 6199.5342 | 6942.32 | 6353.867 | 6145.893 | 13266.08 | 6530.491 | 6242.955 | 7035.835 |
| Std | 0.0029181 | 368.8253 | 418.47407 | 342.971 | 563.4305 | 219.693 | 3334.069 | 399.0299 | 248.7349 | 2850.707 |
| Rank | 1 | 3 | 4 | 9 | 6 | 2 | 10 | 7 | 5 | 8 |

Minimize $f(\vec{x}) = 0.7854x_1x_2^2\left(3.3333x_3^2 + 14.9334x_3 - 43.0934\right) - 1.508x_1\left(x_6^2 + x_7^2\right)$
$+ 7.4777\left(x_6^3 + x_7^3\right)$

Subject to $g_1(\vec{x}) = \frac{27}{x_1x_2^2x_3} - 10$

$g_2(\vec{x}) = \frac{397.5}{x_1x_2^2x_3^2} - 10$

$g_3(\vec{x}) = \frac{1.93x_4^3}{x_2x_3x_6^4} - 10$

$g_4(\vec{x}) = \frac{1.93x_5^3}{x_2x_3x_7^4} - 10$

$g_5(\vec{x}) = \frac{\sqrt{\frac{745x_4}{x_2x_3}^2 + 16.9\times10^6}}{110.0x_6^3} - 10$

$g_6(\vec{x}) = \frac{\sqrt{\frac{745x_4}{x_2x_3}^2 + 157.5\times10^6}}{85.0x_6^3} - 10$

$g_7(\vec{x}) = \frac{x_2x_3}{40} - 10$

$g_8(\vec{x}) = \frac{5x_2}{x_1} - 10$

$g_9(\vec{x}) = \frac{x_1}{12x_2} - 10$

$g_{10}(\vec{x}) = \frac{1.5x_6 + 1.9}{x_4} - 10$

$g_{11}(\vec{x}) = \frac{1.1x_7 + 1.9}{x_5} - 10$

Variables range $2.6x13.6$
$0.7x20.8$
$17x328$
$7.3x48.3$
$7.8x58.3$
$2.9x63.9$
$5.0x75.5$

(23)

The speed reducer engineering problem was tackled using the proposed mSHO algorithm and other competitive algorithms, as depicted in Table 8. The obtained statistical results are presented in Table 9. As demonstrated in Table 9, the optimal value of the function is 2993.634, which was achieved using the mSHO, HGS, AOA, and PSO algorithms. The results reveal that the mSHO algorithm produces promising outcomes compared to the other algorithms and has the potential to achieve minimal total weight for the speed reducer in this problem.

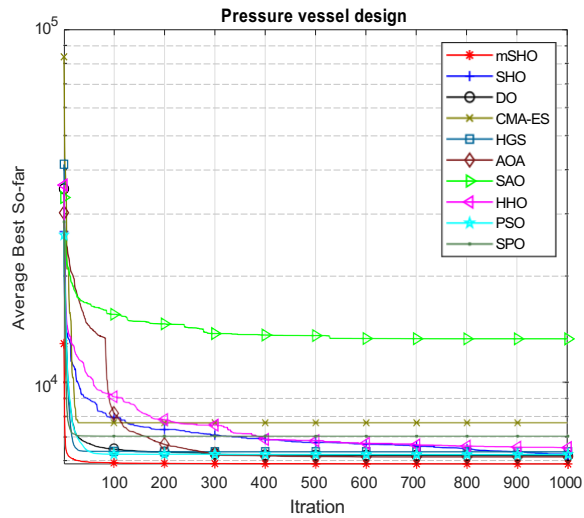

**(a)** Convergence Curve

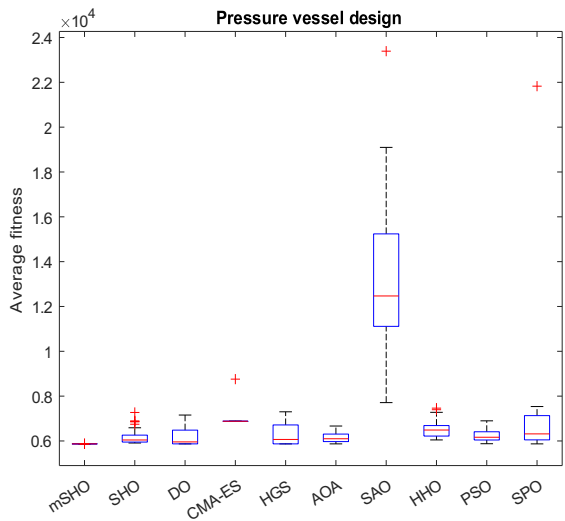

**(b)** Boxplot

**Figure 4:** Convergence curve and Boxplot for mSHO against other competitors – pressure vessel design problem

**Table 8:** Best solution obtained from the comparative algorithms for solving speed reducer design problem

| Algorithm | x1 | x2 | x3 | x4 | x5 | x6 | x7 | Cost |
|---|---|---|---|---|---|---|---|---|
| mSHO | 3.497599 | 0.7 | 17 | 7.3 | 7.713535 | 3.350056 | 5.285631 | 2993.634 |
| SHO | 3.498576 | 0.7 | 17 | 7.3 | 7.70853 | 3.349319 | 5.28455 | 2994.77 |
| DO | 3.497563 | 0.7 | 17 | 7.300001 | 7.713536 | 3.350059 | 5.285605 | 2993.635 |
| CMA-ES | 3.6 | 0.7 | 17 | 7.3 | 8.072148 | 3.402879 | 5.312241 | 3071.526 |
| HGS | 3.497599 | 0.7 | 17 | 7.3 | 7.713535 | 3.350056 | 5.285631 | 2993.634 |
| AOA | 3.497599 | 0.7 | 17 | 7.3 | 7.713535 | 3.350056 | 5.285631 | 2993.634 |
| SAO | 3.6 | 2.6 | 3.565453 | 2.84373 | 2.951027 | 3.314187 | 2.715053 | 3230.902 |
| HHO | 3.49797 | 0.7 | 17 | 7.3 | 7.732423 | 3.350521 | 5.285312 | 2994.197 |
| PSO | 3.497599 | 0.7 | 17 | 7.3 | 7.713535 | 3.350056 | 5.285631 | 2993.634 |
| SPO | 3.497613 | 0.7 | 17 | 7.3 | 7.713331 | 3.350886 | 5.285453 | 2993.836 |

Figure 5 depicts the convergence curves and boxplot for the mSHO algorithm and other compared methods for the speed reducer design problem. As shown in the figure, the mSHO algorithm converges faster than the other algorithms and can usually obtain the near-optimal solution more rapidly. Although the other algorithms also showed competitive performance, the HHO and SAO had the lowest performance. On the other hand, the boxplot results illustrate the stability of the proposed mSHO algorithm, followed by the HGS and PSO algorithms. Overall, the experiment's findings reveal the efficacy and stability of the proposed mSHO algorithm in tackling the speed reducer design problem.

**Table 9:** Results obtained from competitor algorithms for speed reducer engineering problem

| Mea. | mSHO | SHO | DO | CMA-ES | HGS | AOA | SAO | HHO | PSO | SPO |
|------|------|-----|-----|--------|-----|-----|-----|-----|-----|-----|
| Min | 2993.6343 | 2994.7697 | 2993.63462 | 3071.526 | 2993.634 | 2993.634 | 3230.902 | 2994.197 | 2993.634 | 2993.836 |
| Max | 2993.6343 | 3821.162 | 3001.26628 | 3072.334 | 2993.717 | 3006.627 | 8071.254 | 4307.676 | 2993.634 | 3359.51 |
| Mean | 2993.6343 | 3139.9316 | 2994.73651 | 3072.307 | 2993.637 | 2994.509 | 4153.676 | 3146.587 | 2993.634 | 3118.58 |
| Std | 2.67E-13 | 200.19962 | 1.80047494 | 0.147413 | 0.015026 | 2.925884 | 956.0172 | 259.7566 | 2.67E-13 | 107.09919 |
| Rank | 1 | 8 | 5 | 7 | 3 | 4 | 10 | 9 | 2 | 6 |

### 6.3 Tension/compression spring problem

The Tension/Compression Spring Design optimization Problem, is a mechanical engineering problem that aims to minimize the weight of the spring while ensuring that certain constraints are satisfied (Bhadoria & Kamboj, 2019). The problem involves selecting the optimal values for parameters such as wire diameter ($d$), number of active coils ($N$), and mean coil diameter ($D$). Constraints are placed on the surge frequency, minimum deflection, and shear stress. The goal is to find the optimal combination of parameters that satisfies all constraints while minimizing the weight of the spring. The following equations present the mathematical model for this particular engineering design problem.

Consider $\vec{x} = \begin{bmatrix} x_1 & x_2 & x_3 \end{bmatrix} = \begin{bmatrix} d & D & N \end{bmatrix}$

Minimize $f(\vec{x}) = (x_3 + 2) x_2 x_1^2$

Subject to $\quad g_1(\vec{x}) = 1 - \dfrac{x_2^3 x_3}{71785 x_1^4} \leq 0$

$$g_2(\vec{x}) = \dfrac{4x_2^2 - x_1 x_2}{12566\left(x_2 x_1^3 - x_1^4\right)} + \dfrac{1}{5108 x_1^2} \leq 0 \qquad (24)$$

$g_3(\vec{x}) = 1 - \dfrac{140.45 x_1}{x_2^2 x_3} \leq 0$

$g_4(\vec{x}) = \dfrac{x_1 + x_2}{1.5} - 1 \leq 0$

Variables range $\quad 0.05 \leq x_1 \leq 2$
$0.25 \leq x_2 \leq 1.30$
$2.00 \leq x_3 \leq 15$

The proposed mSHO algorithm and other competitive algorithms were employed to solve this engineering problem as presented in Table 10. The statistical results obtained from the experiments are provided in Table 11. It is evident from Table 11 that the mSHO algorithm achieved the optimal value of the function, which was 0.012665. The results indicate that the mSHO algorithm performed significantly better than the other algorithms in achieving a minimal weight of the tension spring in this problem.

Figure 6 presents the convergence curves and boxplot for the Tension/Compression Spring Design problem solved using the mSHO algorithm and other competitive methods. The results show that the proposed mSHO algorithm outperforms the other algorithms, achieving the near-optimal solution faster. The SAO and HGS algorithms exhibit the lowest performance, while the AOA and SHO algorithms perform competitively. The boxplot analysis further confirms the stability of the mSHO algorithm, followed by AOA and SHO. These results demonstrate the efficiency and stability of the mSHO algorithm in solving the Tension/Compression Spring Design problem.

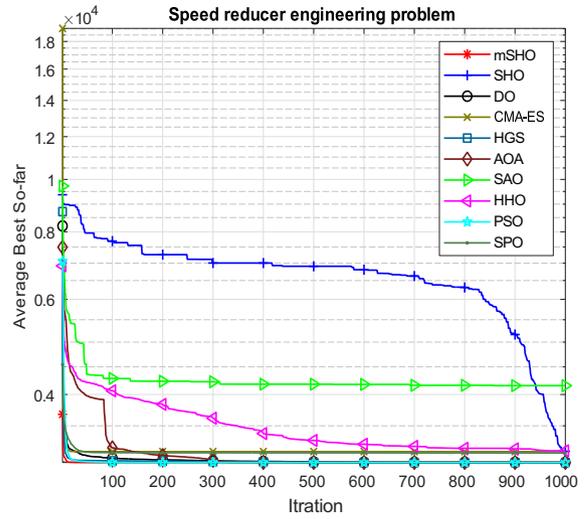

**(a)** Convergence Curve

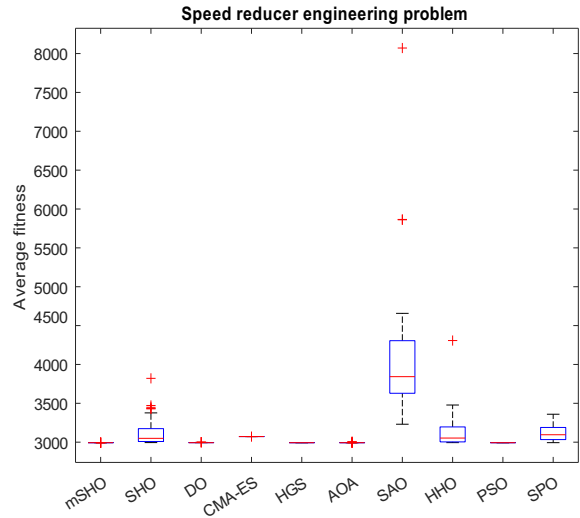

**(b)** Boxplot

**Figure 5:** Convergence curve and Boxplot for mSHO against other competitors – Speed reducer engineering problem

**Table 10:** Best solution obtained from the comparative algorithms for solving tension/compression spring problem

| Algorithm | x1 | x2 | x3 | Cost |
|-----------|-----|-----|-----|------|
| mSHO | 0.051687 | 0.356672 | 11.29167 | 0.012665 |
| SHO | 0.050987 | 0.340068 | 12.33627 | 0.012674 |
| DO | 0.051983 | 0.363837 | 10.88351 | 0.012667 |
| CMA-ES | 0.05 | 0.311342 | 15 | 0.013232 |
| HGS | 0.05251 | 0.376782 | 10.20369 | 0.012678 |
| AOA | 0.051803 | 0.359477 | 11.12903 | 0.012665 |
| SAO | 1.227483 | 1.068724 | 0.631804 | 0.013213 |
| HHO | 0.052513 | 0.376854 | 10.19919 | 0.012677 |
| PSO | 0.051671 | 0.356288 | 11.31423 | 0.012665 |
| SPO | 0.051436 | 0.350646 | 11.65453 | 0.012667 |

**Table 11:** Results obtained from competitor algorithms for tension/compression spring problem

| Mea. | mSHO | SHO | DO | CMA-ES | HGS | AOA | SAO | HHO | PSO | SPO |
|------|------|-----|-----|--------|-----|-----|-----|-----|-----|-----|
| Min | 0.012665 | 0.012674 | 0.012667 | 0.013232 | 0.012678 | 0.012665 | 0.013213 | 0.012677 | 0.012665 | 0.012667 |
| Max | 0.012738 | 0.014513 | 0.014664 | 0.013278 | 1373.172 | 0.014318 | 0.029041 | 0.016977 | 0.016907 | 0.030455 |
| Mean | 0.012672 | 0.012965 | 0.013254 | 0.013277 | 45.79453 | 0.013044 | 0.020018 | 0.013592 | 0.013508 | 0.014547 |
| Std | 1.44E-05 | 0.000415 | 0.000563 | 8.44E-06 | 250.7016 | 0.000507 | 0.003995 | 0.000955 | 0.001161 | 0.004475 |
| Rank | 1 | 2 | 4 | 5 | 10 | 3 | 9 | 7 | 6 | 8 |

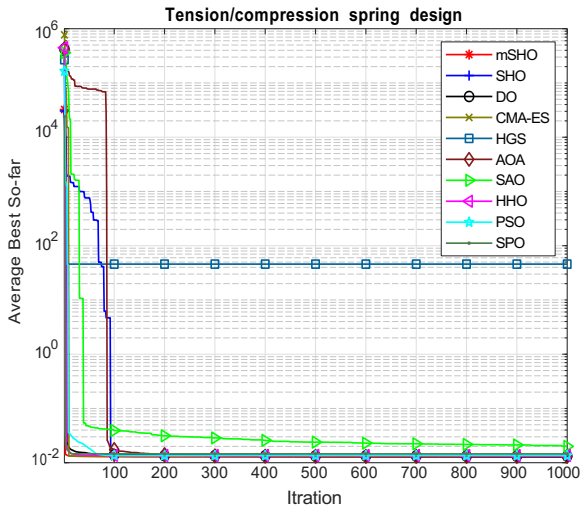

**(a)** Convergence Curve

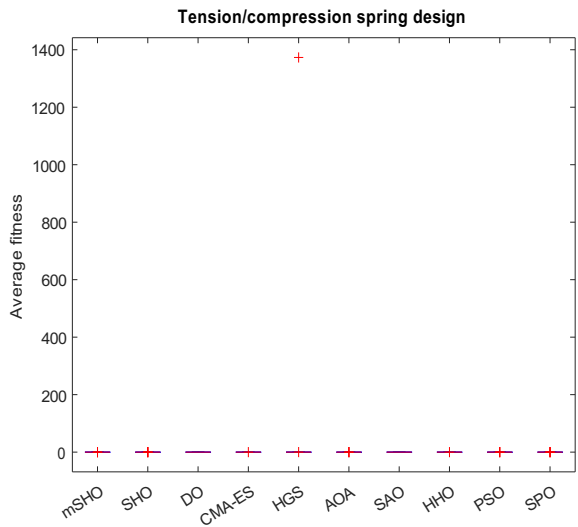

**(b)** Boxplot

**Figure 6:** Convergence curve and Boxplot for mSHO against other competitors – Tension/compression spring problem

### 6.4 Welded beam design problem

The Welded Beam Design Problem is another important engineering design problem that has been considered in previous research Sadollah et al., 2013. The main objective of this problem is to minimize the cost of fabricating a welded beam by optimizing four variables: bar thickness (b), bar length including attached parts (l), weld thickness (h), and bar height (h). The problem is subject to four constraints, including buckling constraints of the bar (Pc), side constraints, end deflection of the beam (d), bending stress of the beam (h), and shear stress. The mathematical model for this problem is given below.

$$
\begin{aligned}
&\text{Consider} && \vec{x} = [x_1 x_2 x_3 x_4] = [\text{hlttbb}] \\
&\text{Minimize} && f(\vec{x}) = 1.10471 x_1^2 x_2 + 0.04811 x_3 x_4 (14.0 + x_2) \\
&\text{Subject to} && g_1(\vec{x}) = \tau(\vec{x}) - \tau_{max} \leq 0 \\
& && g_2(\vec{x}) = \sigma(\vec{x}) - \sigma_{max} \leq 0 \\
& && g_3(\vec{x}) = \delta(\vec{x}) - \delta_{max} \leq 0 \\
& && g_4(\vec{x}) = x_1 - x_4 \leq 0 \\
& && g_5(\vec{x}) = P - P_c(\vec{x}) \leq 0 \\
& && g_6(\vec{x}) = 0.125 - x_1 \leq 0 \\
& && g_7(\vec{x}) = 1.10471 x_1^2 + 0.04811 x_3 x_4 (14.0 + x_2) - 5.00 \leq 0
\end{aligned}
$$

$$
\begin{aligned}
&\text{Variables range} && 0.1 \leq x_1 \leq 2 \\
& && 0.1 \leq x_2 \leq 10 \\
& && 0.1 \leq x_3 \leq 10 \\
& && 0.1 \leq x_4 \leq 2
\end{aligned}
$$

$$
\begin{aligned}
&\text{where} && \tau(\vec{x}) = \sqrt{(\tau')^2 + 2\tau'\tau''\frac{x_2}{2R} + (\tau'')^2}, \\
& && \tau' = \sqrt{\frac{P}{2 x_1 x_2}}, \tau'' = \frac{MR}{J} \\
& && M = P\left(L + \frac{x_2}{2}\right) \\
& && R = \sqrt{\frac{x_2^2}{4} + \left(\frac{x_1 + x_3}{2}\right)^2} \\
& && J = 2\left\{\sqrt{2 x_1 x_2}\left[\frac{x_2^2}{4} + \left(\frac{x_1 + x_3}{2}\right)^2\right]\right\} \\
& && \sigma(\vec{x}) = \frac{6PL}{x_4 x_3^2}, \quad \delta(\vec{x}) = \frac{6PL^3}{Ex_3^3 x_4} \\
& && P_c(\vec{x}) = \frac{4.013 E \sqrt{\frac{x_3^2 x_6^6}{36}}}{L^2}\left(1 - \frac{x_3}{2L}\sqrt{\frac{E}{4G}}\right) \\
& && P = 6000 \text{lb}, L = 14 \text{in.}, \quad \delta_{max} = 0.25 \text{ in.} \\
& && E = 30 \times 1^6 \text{psi}, \quad G = 12 \times 10^6 \text{psi} \\
& && \tau_{max} = 13600 \text{psi}, \quad \sigma_{max} = 30000 \text{psi}
\end{aligned}
$$
(25)

This engineering problem is solved using the proposed mSHO and other competitive algorithms as shown in Table 12. The obtained statistical results are presented in Table 13. Table 13 shows that the optimal value of the function is 1.724967, which was achieved using the mSHO algorithm. As the results show, mSHO produces promising results in comparison with the other algorithms and has a good ability for achieving minimal fabrication's cost in this problem.

The performance of the mSHO algorithm and other competitive algorithms in solving the Welded Beam Design Problem is depicted in Figure 7. The results show that the mSHO algorithm converges faster than the other algorithms and achieves near-optimal solutions more quicker. Although the other algorithms also perform competitively, the SAO and CMA-ES algorithms show the lowest performance. Furthermore, the boxplot results demonstrate the stability of the mSHO algorithm, followed by the DO and AOA algorithms. These results indicate the efficiency and stability of the mSHO algorithm in solving the Welded Beam Design Problem.

**Table 12:** Best solution obtained from the comparative algorithms for solving welded beam design problem

| Algorithm | x1 | x2 | x3 | x4 | Cost |
|---|---|---|---|---|---|
| mSHO | 0.20573 | 3.470471 | 9.036627 | 0.20573 | 1.724852 |
| SHO | 0.192302 | 3.778244 | 9.052142 | 0.205653 | 1.746601 |
| DO | 0.20573 | 3.470495 | 9.036626 | 0.20573 | 1.724854 |
| CMA-ES | 0.205259 | 3.492761 | 9.043531 | 0.205725 | 1.728297 |
| HGS | 0.205736 | 3.470414 | 9.036457 | 0.205737 | 1.724881 |
| AOA | 0.20573 | 3.470473 | 9.036624 | 0.20573 | 1.724852 |
| SAO | 2 | 0.932642 | 1.496869 | 0.644415 | 1.792207 |
| HHO | 0.173239 | 4.308508 | 9.096589 | 0.205631 | 1.790459 |
| PSO | 0.20573 | 3.470475 | 9.036624 | 0.20573 | 1.724852 |
| SPO | 0.20573 | 3.470484 | 9.036623 | 0.20573 | 1.724852 |

**Table 13:** Results obtained from competitor algorithms for the welded beam problem

| Mea. | mSHO | SHO | DO | CMA-ES | HGS | AOA | SAO | HHO | PSO | SPO |
|---|---|---|---|---|---|---|---|---|---|---|
| Min | 1.724852 | 1.746601 | 1.724854 | 1.728297 | 1.724881 | 1.724852 | 1.792207 | 1.790459 | 1.724852 | 1.724852 |
| Max | 1.7256 | 4.306289 | 1.744596 | 2.266116 | 4.341756 | 1.836955 | 4.036102 | 2.682967 | 2.175585 | 3.03395 |
| Mean | 1.724967 | 1.995641 | 1.72778 | 2.230261 | 2.13089 | 1.733932 | 2.995645 | 2.025109 | 1.814491 | 2.067492 |
| Std | 0.000192 | 0.477618 | 0.00427 | 0.136449 | 0.631101 | 0.024171 | 0.596869 | 0.210771 | 0.13514 | 0.387053 |
| Rank | 1 | 5 | 2 | 9 | 8 | 3 | 10 | 6 | 4 | 7 |

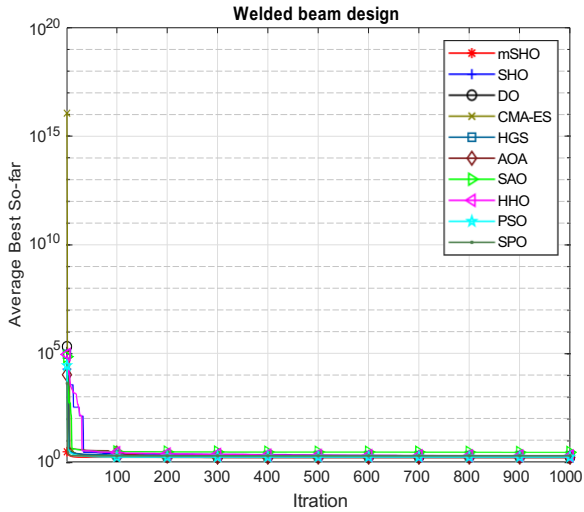

**(a)** Convergence Curve

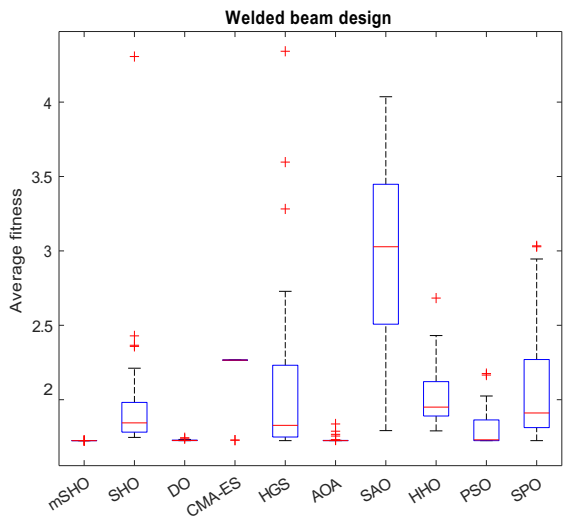

**(b)** Boxplot

**Figure 7:** Convergence curve and Boxplot for mSHO against other competitors – Welded beam design problem

## 6.5 Three-bar truss engineering design problem

The aim of this engineering design problem is to minimize the weight of a truss by optimizing two parameters that represent the cross-sectional areas ($x_1$ and $x_2$), subject to the bounds constraints of $0 \le x_1, x_2 \le 1$. Additionally, three inequality constraints are related to buckling, deflection, and stress. The mathematical representation of this problem is as follows:

$$\text{Consider } \vec{x} = [x_1 \quad x_2] = [A_1 \quad A_2]$$
$$\text{Minimize } f(\vec{x}) = \left( 2\sqrt{2x_1} + x_2 \right) * l,$$
$$\text{Subject to } g_1(\vec{x}) = \frac{2x_1 + x_2}{\sqrt{2x_1^2 + 2x_1 x_2}} P - \sigma 0$$
$$g_2(\vec{x}) = \frac{x_2}{\sqrt{2x_1^2 + 2x_1 x_2}} P - \sigma 0$$
$$g_3(\vec{x}) = \frac{1}{\sqrt{2x_1^2 + 2x_1}} P - \sigma 0 \qquad (26)$$
$$\text{Variables range} \quad 0 \, x_1, x_2 \, 1$$
$$\text{where} \quad l = 100 \text{ cm}, P = \frac{2\text{KN}}{\text{cm}^2}, \sigma = \frac{2\text{KN}}{\text{cm}^2}$$

The engineering problem is tackled using the proposed mSHO algorithm and other competitive algorithms, as shown in Table 14. The statistical results obtained are presented in Table 15, which indicates that the mSHO algorithm achieved the optimal value of the function, 263.8915. The results demonstrate that mSHO produces better results than the other algorithms and has a strong ability to minimize the weight of the truss in this problem.

In addition, Figure 8 displays the convergence curves and box-plot for the mSHO and other compared algorithms in solving the three-bar truss design problem. The figure demonstrates that the mSHO algorithm converges faster than the other methods and can generally achieve near-optimal solutions more quickly. Although the other algorithms also exhibit competitive performance, the SAO and HGS show the lowest performance. Moreover, the box-plot results indicate the stability of the mSHO algorithm, followed by the PSO and AOA algorithms. These results suggest that the proposed mSHO algorithm is efficient and stable in solving the three-bar truss design problem.

**Table 14:** Best solution obtained from the comparative algorithms for solving three-bar truss engineering design problem

| Algorithm | x1 | x2 | Cost |
|---|---|---|---|
| mSHO | 0.788649 | 0.408235 | 263.8915 |
| SHO | 0.787638 | 0.411102 | 263.8922 |
| DO | 0.788649 | 0.408234 | 263.8915 |
| CMA-ES | 0.756483 | 0.508411 | 264.8067 |
| HGS | 0.780942 | 0.431176 | 264.0014 |
| AOA | 0.788649 | 0.408235 | 263.8915 |
| SAO | 1 | 0.733032 | 264.4989 |
| HHO | 0.788486 | 0.408697 | 263.8915 |
| PSO | 0.788649 | 0.408235 | 263.8915 |
| SPO | 0.788651 | 0.408229 | 263.8915 |

**Table 15:** Results obtained from competitor algorithms for three-bar truss engineering design problem

| Mea. | mSHO | SHO | DO | CMA-ES | HGS | AOA | SAO | HHO | PSO | SPO |
|---|---|---|---|---|---|---|---|---|---|---|
| Min | 263.8915 | 263.8922 | 263.8915 | 264.8067 | 264.0014 | 263.8915 | 264.4989 | 263.8915 | 263.8915 | 263.8915 |
| Max | 263.8915 | 264.2569 | 263.8919 | 265.1765 | 289.6074 | 263.8915 | 308.1042 | 264.4841 | 263.8915 | 269.4398 |
| Mean | 263.8915 | 263.9661 | 263.8915 | 264.819 | 269.3437 | 263.8915 | 274.0035 | 264.001 | 263.8915 | 264.0988 |
| Std | 2.59E-11 | 0.086065 | 8.03E-05 | 0.067512 | 4.670943 | 7.45E-07 | 9.827149 | 0.128725 | 3.61E-07 | 1.011145 |
| Rank | 1 | 5 | 4 | 8 | 9 | 3 | 10 | 6 | 2 | 7 |

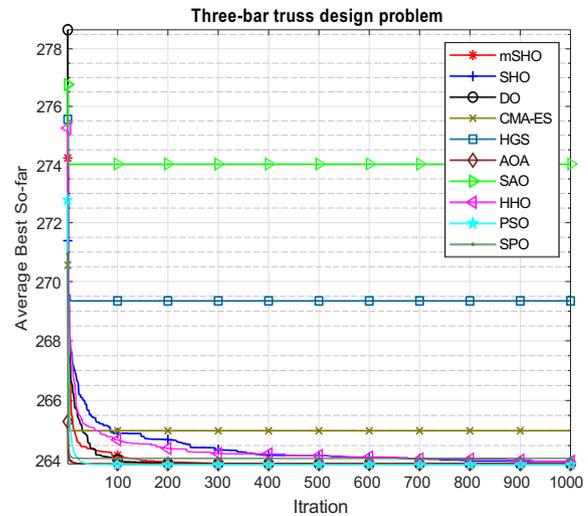

**(a)** Convergence Curve

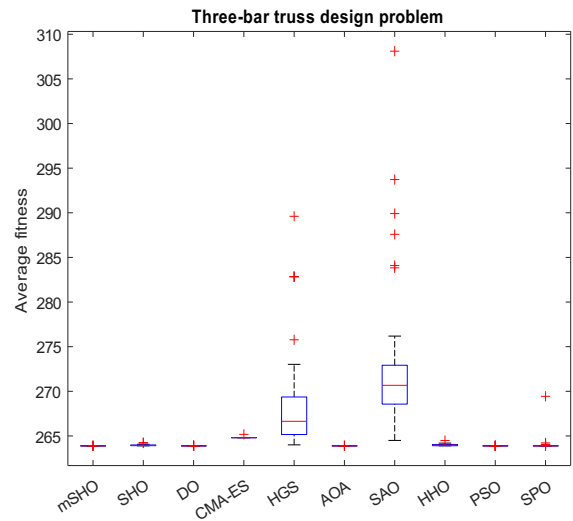

**(b)** Boxplot

**Figure 8:** Convergence curve and Boxplot for mSHO against other competitors – Three-bar truss engineering design problem

## 6.6 Industrial refrigeration system problem

The objective of the industrial refrigeration system problem is to minimize the cost of the refrigeration system while optimizing the refrigerants, temperature levels, cycle configuration, and compression technology. This problem is described mathematically and has multiple variables and constraints. The details of the problem formulation can be found in (Marechal & Kalitventzeff, 2001).

The problem can be mathematically formulated as follows:

Minimize
$$f(x) = 63098.88x_2x_4x_{12} + 5441.5x_2^2x_{12} + 115055.5x_2^{1.664}x_6$$
$$+6172.27x_2^2x_6 + 63098.88x_1x_4x_9 + 5441.5x_1^2x_9 + 115055.5x_1^{1.664}x_5$$
$$+6172.27x_1^2x_5 + 140.53x_1x_{11} + 281.29x_3x_{11} + 70.26x_2^2 + 281.29x_3x_{10}$$
$$+281.29x_3^2x_1 + 14437x_8^{1.8812}x_{12}^{0.3424}x_{10}x_{14}^{-1}x_1^2x_7x_9^{-1} + 20470.2x_3^2x_{10}$$

Subject to
$$g_1(x) = 1.524x_4^{-1} \le 1$$
$$g_2(x) = 1.524x_5^{-1} \le 1$$
$$g_3(x) = (x_8 - 2x^2 \cdot x_7 - 1 \le 0)$$
$$g_4(x) = 0.07789x_1 - 2x_7^{-1}x_9 - 1 \le 0,$$
$$g_5(x) = 7.05305x_9^{-1}x_1^2x_{10}x_7^{-1}x_8^{-1}x_{14}^{-1}x_1^{-1} - 1 \le 0,$$
$$g_6(x) = 0.0833x_{13}^{-1}x_{14} - 1 \le 0,$$
$$g_7(x) = 47.136x_2^{0.333}x_{10}^{-1}x_{12} - 1.333x_8x_{13}^{2.1195}$$
$$+ 62.08x_{13}^{2.1195}x_{12}^{-0.2}x_{10}^{-1} - 1 \le 0$$
$$g_8(x) = 0.04771x_{10}x_8^{1.8812}x_{12}^{0.3424} - 1 \le 0$$
$$g_9(x) = 0.0488x_9x_7^{1.893}x_{11}^{0.316} - 1 \le 0$$
$$g_{10}(x) = 0.0099x_1x_4^{-1} - 1 \le 0,$$
$$g_{11}(x) = 0.0193x_2x_5^{-1} - 1 \le 0$$
$$g_{12}(x) = 0.0298x_1x_7^{-1} - 1 \le 0$$
$$g_{13}(x) = 0.056x_2x_8^{-1} - 1 \le 0$$
$$g_{14}(x) = 2x_9^{-1} - 1 \le 0,$$
$$g_{15}(x) = 2x_{10}^{-1} - 1 \le 0,$$
$$g_{16}(x) = x_{12}x_{11}^{-1} - 1 \le 0$$

Variables range $\quad 0.001 \le x_i \le 5, i = 1, \cdots, 14$

(27)

The proposed mSHO algorithm and other competitive algorithms are used to solve this engineering problem, as shown in Table 16. The statistical results obtained are presented in Table 17, which indicates that the mSHO algorithm achieved the optimal value of the function at 0.032255. The results demonstrate that mSHO yields promising outcomes compared to other algorithms and is proficient in achieving the minimum cost of the refrigeration system in this problem.

**Table 16:** Best solution obtained from the comparative algorithms for solving industrial refrigeration system problem

| Algorithm | x1 | x2 | x3 | x4 | x5 | x6 | x7 | x8 | x9 | x10 | x11 | x12 | x13 | x14 | Cost |
|---|---|---|---|---|---|---|---|---|---|---|---|---|---|---|---|
| mSHO | 0.001 | 0.001 | 0.001 | 0.001 | 0.001 | 1.524 | 1.224 | 4.999996 | 2 | 2.002 | 0.001 | 0.001 | 0.047379 | 0.632221 | 0.032255 |
| SHO | 0.001 | 0.001 | 0.001 | 0.001 | 0.001 | 1.527144 | 1.22777 | 4.855744 | 1.9948 | 0.001 | 0.001 | 0.001 | 0.001 | 0.060452 | 0.095676 |
| DO | 0.001 | 0.001061 | 0.001638 | 0.001813 | 0.001 | 1.524 | 1.524062 | 4.999996 | 1 | 2.22285 | 1 | 1 | 0.001 | 0.001 | 0.032272 |
| CMA-ES | 0.001 | 0.001 | 0.041677 | 0.001 | 0.001 | 1.524 | 2.222805 | 5 | 1 | 1 | 0.001 | 0.001 | 0.001 | 0.047304 | 4311.659 |
| AOA | 0.001 | 0.001 | 0.001 | 0.001 | 0.001 | 1.524 | 2.611889 | 5 | 1.1 | 1 | 1 | 0.001 | 0.001 | 0.047384 | 0.037211 |
| SAO | 0.001661 | 1.187545 | 1.85308 | 0.162425 | 0.235081 | 4.316108 | 2.01633 | 2.968422 | 4.995771 | 3 | 3.974443 | 4.264568 | 4.244538 | 1 | 14246044 |
| HHO | 0.001 | 0.001 | 0.001993 | 0.226003 | 0.001 | 2.214877 | 2.668151 | 5 | 1 | 1 | 0.001 | 0.001 | 0.046788 | 0.637984 | 0.364219 |
| PSO | 0.001 | 0.001 | 0.001 | 0.001 | 0.001 | 1.524 | 1.224687 | 5 | 2 | 2 | 0.001 | 0.001 | 0.047307 | 0.637507 | 0.032213 |
| SPO | 0.001 | 0.001 | 0.001576 | 0.002884 | 0.001 | 1.524 | 1.225 | 5 | 2 | 2.002 | 0.001 | 0.001 | 0.047 | 0.62 | 0.051055 |

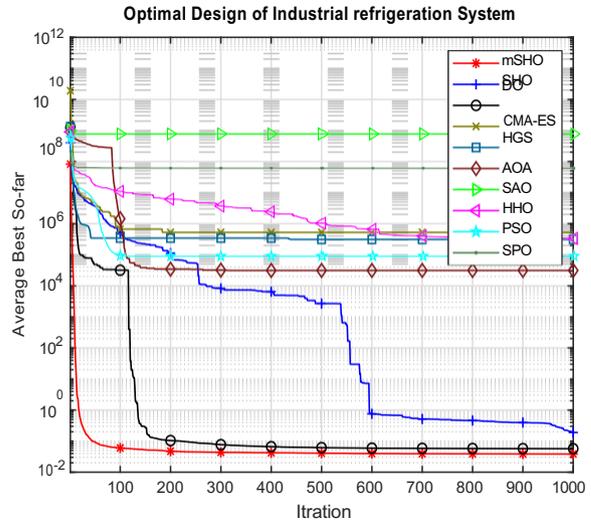

These findings demonstrate the effectiveness and stability of the proposed mSHO algorithm in solving the industrial refrigeration system optimization problem.

**(a)** Convergence Curve

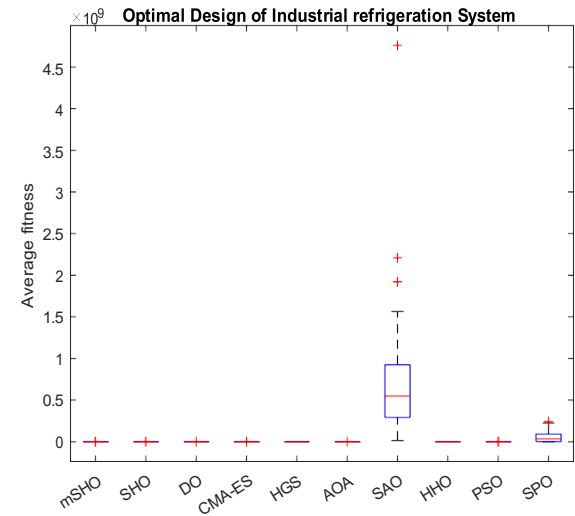

**(b)** Boxplot

**Figure 9:** Convergence curve and Boxplot for mSHO against other competitors – industrial refrigeration system problem

**Table 17:** Results obtained from competitor algorithms for industrial refrigeration system problem

| Mea. | mSHO | SHO | DO | CMA-ES | HGS | AOA | SAO | HHO | PSO | SPO |
|---|---|---|---|---|---|---|---|---|---|---|
| Min | 0.032255 | 0.095676 | 0.032272 | 4311.659 | 0.032213 | 0.032711 | 14246044 | 0.364219 | 0.032213 | 0.058815 |
| Max | 0.052875 | 0.324592 | 0.13298 | 2660163 | 9361894 | 9361894 | 4.76E+09 | 984165.9 | 9361894 | 2.44E+08 |
| Mean | 0.038305 | 0.189575 | 0.056888 | 312063.2 | 31206.37 | 7.88E+08 | 3498335 | 93619 | 93619 | 62388451 |
| Std | 0.005233 | 0.043447 | 0.023447 | 484889.9 | 448868.4 | 170924 | 9.26E+08 | 466964.4 | 2856581 | 74649811 |
| Rank | 1 | 3 | 2 | 8 | 6 | 4 | 10 | 7 | 5 | 9 |

Figure 9 shows the convergence curves and boxplot for the industrial refrigeration system optimization problem using mSHO and the compared algorithms. The results indicate that the mSHO algorithm achieves faster convergence and usually obtains near-optimal solutions quicker than the other algorithms. Although the other algorithms also demonstrate competitive performance, the SAO and SPO algorithms show the lowest performance. Furthermore, the boxplot results show that the proposed mSHO algorithm exhibits stability in comparison to the DO and SHO algorithms.

## 6.7 Multi-Product batch plant problem

The objective of this model is to minimize the production cost of a multi-product batch process by optimizing the allocation of resources. The process consists of three stages that all products follow, and there are two different products being produced. The model has ten decision variables: $N_1, N_2, N_3, V_1, V_2, V_3, T_1, T_2, B_1,$ and $B_2$, represented by the shorthand notations $x_1$ through $x_{10}$. The mathematical formulation of the model, as presented in

**kumar2020test**, is as follows:

Minimize $f(x) = \sum_{j=1}^{M} a_j N_j V_j^{b_j}$

Subject to
$$g_1(x) = S_{ij}B_i - V_j \leq 0$$
$$g_2(x) = -H + \sum_{i=1}^{N} \frac{Q_i T_i}{B_i} \leq 0$$
$$g_3(x) = t_{ij} - N_j T_i \leq 0$$

Variables range
$$1 \leq N_i \leq 3$$
$$250 \leq V_j \leq 2500$$
$$\max\left(\frac{t_{ij}}{N_j}\right) \leq T_i \leq \max\left(t_{ij}\right)$$
$$\frac{Q_i^* T_i}{H} \leq B_i \leq \min\left(Q_i, \min\frac{V_j^u}{S_{ij}}\right)$$

Where
$N = 2, M = 3, a_j = 250, H = 6000, b_j = 0.6, S_{11} = 2,$
$S_{12} = 3, S_{13} = 4, S_{21} = 4, S_{22} = 6, S_{23} = 3, t_{11} = 8, t_{12} = 20$
$t_{13} = 8, t_{21} = 16, t_{22} = 4, t_{23} = 4. i$ means the product, $j$ means stage the stage of production, $a_j$ is variable cost coefficient of stage $j$ process equipment investment cost, $N_j$ is the number of equipment at stage $j$, $V_j$ is the size of equipment at stage $j$, $T_i$ means the cycle time of product $i$, $B_i$ means the batch size of product $i$, $b_j$ is the fixed-cost charges for the investment cost of process equipment at stage $j$.

(28)

The multi-product batch process model presented above aims to reduce production costs by optimizing the allocation of resources in product manufacturing. To solve this problem, the proposed mSHO algorithm and several other competitive algorithms were compared, and the results are presented in Table 18. From the statistical results in Table 19, it can be seen that the mSHO algorithm achieved the optimal value of the function, which was 58507.14. The comparison of the algorithms shows that mSHO outperforms the others and can achieve minimal production costs in this problem.

quickly. Although the other algorithms also perform competitively, SAO and SPO demonstrate the poorest performance. Conversely, the boxplot results indicate the stability of the mSHO algorithm, followed by the DO and PSO algorithms. These findings indicate that the proposed mSHO algorithm is effective and stable in addressing the multi-product batch plant problem.

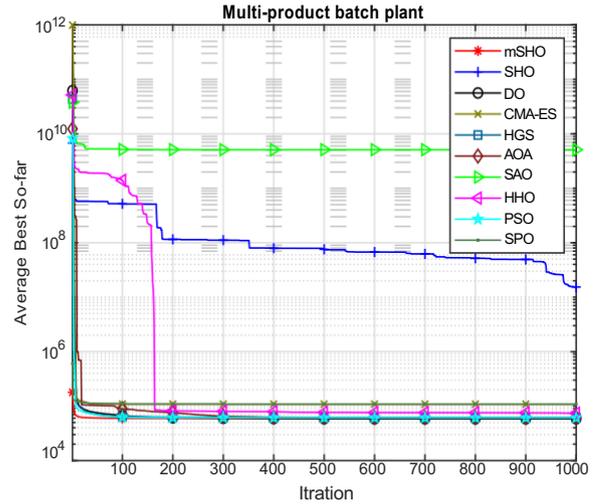

**(a)** Convergence Curve

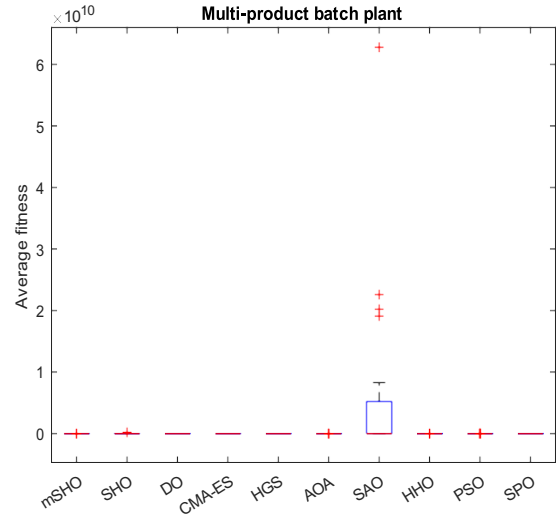

**(b)** Boxplot

**Figure 10:** Convergence curve and Boxplot for mSHO against other competitors – Multi-Product batch plant problem

**Table 18:** Best solution obtained from the comparative algorithms for solving multi-Product batch plant problem

| Algorithms | x1 | x2 | x3 | x4 | x5 | x6 | x7 | x8 | x9 | x10 | Cost |
|---|---|---|---|---|---|---|---|---|---|---|---|
| mSHO | 1.525762 | 1.508602 | 0.674961 | 474.9229 | 719.8871 | 660.2033 | 9.999419 | 7.999732 | 120.1043 | 59.92858 | 58507.14 |
| SHO | 1.519752 | 1.855192 | 0.628977 | 531.0469 | 822.9263 | 705.5837 | 9.992353 | 6.515774 | 120.4098 | 71.54331 | 62676.00 |
| DO | 6720125 | 0.645495 | 1.135445 | 963.3442 | 1445.014 | 1309.242 | 19.99983 | 15.9997 | 234.6931 | 123.4888 | 5363901 |
| CMA-ES | 1.700604 | 2.350703 | 0.549313 | 483.1798 | 735.9209 | 709.1498 | 10.06453 | 8.000826 | 128.8223 | 56.37 | 59471.57 |
| HGS | 651 | 0.730582 | 0.771096 | 980.4283 | 1470.642 | 1286.898 | 20 | 16 | 226.6308 | 134.7916 | 5382053 |
| AOA | 1.4302 | 1.277871 | 0.410608 | 959.619 | 1439.429 | 1321.696 | 20 | 16 | 240.7911 | 119.5092 | 5363.04 |
| SAO | 2.324455 | 2.900101 | 1.435999 | 2.209106 | 1.393715 | 0.792087 | 3.341121 | 1.503093 | 3.441993 | 2.639791 | 80120.89 |
| HHO | 1.713052 | 1.587087 | 1.188423 | 524.3093 | 743.4804 | 1127.821 | 9.999411 | 8.001561 | 150.0532 | 46.29276 | 64778.06 |
| PSO | 1.847141 | 1.97878 | 0.697938 | 479.3873 | 719.881 | 663.0222 | 9.999897 | 7.999933 | 123.3927 | 39.1505 | 58506.03 |
| SPO | 651 | 651 | 651 | 1021.185 | 2088.131 | 1597.147 | 20 | 16 | 332.5038 | 89.04394 | 61399.49 |

**Table 19:** Results obtained from competitor algorithms for multi-Product batch plant problem

| Mea. | mSHO | SHO | DO | CMA-ES | HGS | AOA | SAO | HHO | PSO | SPO |
|---|---|---|---|---|---|---|---|---|---|---|
| Min | 58507.14 | 62676.00 | 53639.01 | 59471.57 | 5382053 | 5363.04 | 80120.89 | 64778.06 | 58506.03 | 61399.49 |
| Max | 66592.15 | 2.12E+08 | 66651.86 | 163121.8 | 7366778 | 66772.51 | 6.28E+10 | 90866.09 | 71888.26 | 1569621 |
| Mean | 58966.85 | 15388990 | 58445.22 | 108483.4 | 6205875 | 6135821 | 5.12E+09 | 74808.67 | 6312671 | 100824.3 |
| Std | 1564.308 | 3E309104 | 4082.766 | 31395.86 | 5715.205 | 2877.974 | 1.24E+10 | 6004.125 | 4792.113 | 28836.15 |
| Rank | 2 | 9 | 1 | 7 | 5 | 4 | 10 | 6 | 3 | 8 |

Furthermore, Figure 10 displays the convergence curves and boxplot for the mSHO algorithm and other competitive methods in solving the multi-product batch plant problem. The graph shows that the proposed mSHO algorithm converges faster than the other algorithms and can typically obtain near-optimal solutions more

### 6.8 Cantilever beam problem

The Cantilever beam problem belongs to the category of concrete engineering problems, as described in the study by Bhadoria et al. (Bhadoria & Kamboj, 2019). The problem aims to minimize the overall weight of a cantilever beam by optimizing the parameters of a hollow square cross-section. The mathematical formulation of this problem is presented below:

Consider $\vec{x} = [x_1 x_2 x_3 x_4 x_5]$

Minimize $f(\vec{x}) = 0.6224 (x_1 + x_2 + x_3 + x_4 + x_5)$

Subject to $g(\vec{x}) = \dfrac{61}{x_1^3} + \dfrac{37}{x_2^3} + \dfrac{19}{x_3^3} + \dfrac{7}{x_4^3} + \dfrac{1}{x_5^3} \le 1$  (29)

Variable range $0.01 \le x_1, x_2, x_3, x_4, x_5 \le 100$

The proposed mSHO algorithm and other competitive algorithms were employed to solve the cantilever beam problem, as shown in Table 20. The statistical results obtained are presented in Table 21, which shows that the optimal value of the function is 1.339956, achieved by using the mSHO algorithm. These results demonstrate that mSHO produces promising outcomes in comparison to the other algorithms, with a high ability to minimize the weight of the cantilever beam in this problem.

**Table 20:** Best solution obtained from the comparative algorithms for solving cantilever beam problem

| Algorithm | x1 | x2 | x3 | x4 | x5 | Cost |
|---|---|---|---|---|---|---|
| mSHO | 6.015906 | 5.308734 | 4.495939 | 3.500899 | 2.152182 | 1.339956 |
| SHO | 6.06771 | 5.389926 | 4.417899 | 3.450059 | 2.155511 | 1.340484 |
| DO | 6.014985 | 5.305781 | 4.490804 | 3.509239 | 2.152893 | 1.339959 |
| CMA-ES | 6.018345 | 5.329495 | 4.474012 | 3.518808 | 2.133878 | 1.340011 |
| HGS | 6.019407 | 5.297822 | 4.482162 | 3.517852 | 2.1567 | 1.339959 |
| AOA | 6.011535 | 5.310885 | 4.495655 | 3.502741 | 2.152851 | 1.339957 |
| SAO | 22.51819 | 66.97232 | 48.25931 | 7.342964 | 15.97925 | 1.558312 |
| HHO | 6.045302 | 5.265671 | 4.576413 | 3.520634 | 2.075623 | 1.340579 |
| PSO | 6.015424 | 5.30515 | 4.496319 | 3.504543 | 2.152239 | 1.339957 |
| SPO | 6.022957 | 5.276079 | 4.519865 | 3.514974 | 2.140741 | 1.340016 |

**Table 21:** Results obtained from competitor algorithms for cantilever beam problem

| Mea. | mSHO | SHO | DO | CMA-ES | HGS | AOA | SAO | HHO | PSO | SPO |
|---|---|---|---|---|---|---|---|---|---|---|
| Min | 1.339956 | 1.340484 | 1.339959 | 1.340011 | 1.339974 | 1.339957 | 1.558312 | 1.340579 | 1.339957 | 1.340016 |
| Max | 1.339967 | 1.351724 | 1.340016 | 1.340075 | 1.34073 | 1.340113 | 10.84034 | 1.344869 | 1.340026 | 3.082532 |
| Mean | 1.339957 | 1.344196 | 1.339972 | 1.340026 | 1.340229 | 1.339991 | 6.036222 | 1.342705 | 1.339974 | 1.552187 |
| Std | 2.08E-06 | 0.0030087 | 1.39E-05 | 1.17E-05 | 0.000202 | 4.44E-05 | 2.312212 | 0.001409 | 1.93E-05 | 0.319524 |
| Rank | 1 | 8 | 2 | 5 | 6 | 4 | 10 | 7 | 3 | 9 |

Additionally, Figure 11 presents the convergence curves and boxplot of mSHO and other compared methods for the cantilever beam problem. The figure indicates that the proposed mSHO algorithm exhibits faster convergence than the other algorithms and can usually obtain near-optimal solutions more quickly. Although the other algorithms also demonstrate competitive performance, SAO and SPO exhibit the lowest performance. Furthermore, the boxplot results reveal the stability of the proposed mSHO algorithm, followed by the DO and PSO algorithms. The results of this experiment demonstrate the efficiency and stability of the proposed mSHO algorithm in solving the cantilever beam problem.

## 6.9 Multiple disc clutch brake problem

The multi-plate disc clutch brake is a well-known optimization problem in mechanical engineering, which aims to minimize the total weight of a multiple-disc clutch brake by optimizing five variables: driving force ($F$), the number of friction surfaces ($Z$),

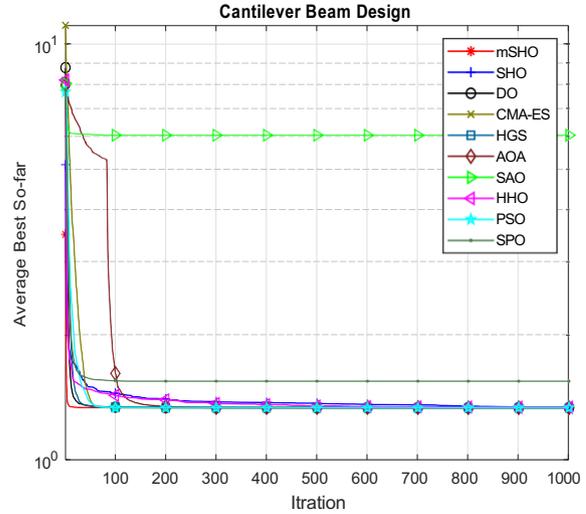

**(a)** Convergence Curve

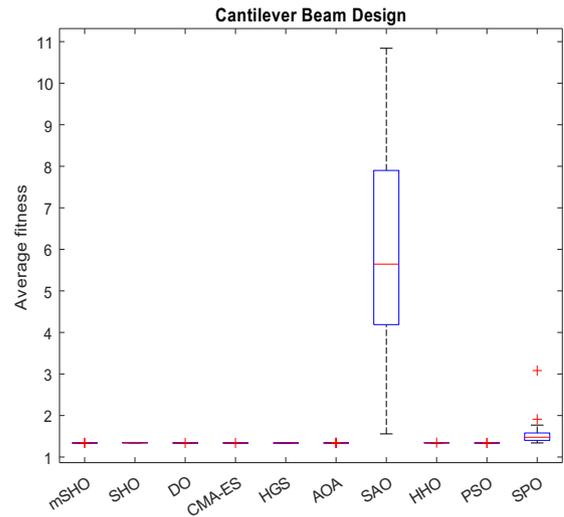

**(b)** Boxplot

**Figure 11:** Convergence curve and Boxplot for mSHO against other competitors – Cantilever beam problem

the thickness of discs ($A$), outer radius ($r0$), and inner radius ($r1$). These variables are denoted by x1, x2, x3, x4, and x5. The problem is subject to eight constraints based on the geometry and operating requirements. The mathematical formulation for this engineering optimization problem can be expressed as follows, as stated in **abderazek2017adaptive**:

Minimize $f(x) = \pi \left( r_0^2 - r_i^2 \right)(Z + 1)\rho t$

Subject to
$$g_1(x) = r_0 - r_i - \Delta r0$$
$$g_2(x) = l_m a x - (Z + 1)(t + \delta)0$$
$$g_3(x) = P_m a x - P_{rz}0$$
$$g_4(x) = P_m a x v_{vr\,max} - P_{rz}v_{sr}0$$
$$g_5(x) = v_{sr\,max} - v_{sr}0$$
$$g_6(x) = T_m a x - T0$$
$$g_7(x) = M_h - sM_s0$$
$$g_8(x) = T0$$

Where
$$M_h = \frac{2}{3}\mu F Z \frac{r_0^3 - r_i^3}{r_0^2 - r_i^2}$$
$$P_{rz} = \frac{2}{3}\pi \frac{F}{\left(r_0^2 - r_i^2\right)}$$
$$vrz = \frac{2\pi n \left(r_0^3 - r_i^3\right)}{90 \left(r_0^3 - r_i^3\right)}$$
$$T = \frac{I_z \pi n}{30\left(M_h - M_f\right)}$$

$\Delta r = 20$ mm, $I_z = 55 \text{kgm}^2$, $P_{max} = 1\text{MPa}$
$F_{max} = 1000$ N, $T_{max} = 15$ s, $\mu = 0.5$
$s = 1.5, M_s = 40\text{Nm}, M_f = 3\text{Nm}, N = 250\text{r/min}$
$v_{sr\,max} = 10$ m/s, $l_{max} = 30$ mm
$60 \text{ mm} \le r_i \le 80 \text{ mm}, 90 \text{ mm} \le r_0 \le, 110 \text{ mm}, 1.5 \text{ mm} \le t \le$
$600N \le F \le 1000N, 2 \le Z \le 9$

(30)

The multi-plate disc clutch brake problem was solved by applying the mSHO algorithm and other competitive algorithms, as presented in Table 22. The statistical analysis of the results is shown in Table 23, which indicates that the mSHO algorithm, along with the AOA algorithm, achieved the minimum objective function value of 0.235242. These results demonstrate that the mSHO algorithm performs better than other algorithms for minimizing the weight of the clutch brake in this engineering problem.

**Table 22:** Results obtained from competitor algorithms for multidisc clutch brake problem

| Algorithm | x1 | x2 | x3 | x4 | x5 | cost |
|-----------|-----|-----|---------|----------|---------|----------|
| mSHO | 70 | 90 | 1 | 213.5391 | 2 | 0.235242 |
| SHO | 69.99874 | 90 | 1 | 66.50295 | 2 | 0.235255 |
| DO | 70 | 90 | 1 | 999.9588 | 2 | 0.235242 |
| CMA-ES | 69.19936 | 90 | 1 | 664.1618 | 2 | 0.243435 |
| HGS | 70 | 90 | 1 | 1000 | 2 | 0.235242 |
| AOA | 70 | 90 | 1 | 858.5333 | 2 | 0.235242 |
| SAO | 64.55532 | 62.9971 | 60.45108 | 61.2192 | 71.05524 | 0.257118 |
| HHO | 70 | 90 | 1 | 946.8728 | 2 | 0.235242 |
| PSO | 70 | 90 | 1 | 33.07846 | 2 | 0.235242 |
| SPO | 70 | 90 | 1 | 1000 | 2 | 0.235242 |

Figure 12 shows the convergence curves and boxplot for mSHO and all other compared methods, revealing that the proposed mSHO algorithm converged faster than the other algorithms and was able to obtain near-optimal solutions more quickly. While the other algorithms also showed competitive performance, the SAO and SPO algorithms exhibited the lowest performance. Additionally, the results of the boxplot demonstrated the stability of the proposed mSHO algorithm, followed by the DO and PSO algorithms. Overall, these findings indicate the efficiency and stability of the mSHO algorithm in handling the multi-plate disc clutch brake problem.

**Table 23:** Best solution obtained from the comparative algorithms for solving multidisc clutch brake problem

| Mea. | mSHO | SHO | DO | CMA-ES | HGS | AOA | SAO | HHO | PSO | SPO |
|------|------|-----|-----|--------|-----|-----|-----|-----|-----|-----|
| Min | 0.235242 | 0.235255 | 0.235242 | 0.243435 | 0.235242 | 0.235242 | 0.257118 | 0.235242 | 0.235242 | 0.235242 |
| Max | 0.235242 | 0.237529 | 0.235242 | 0.24762 | 0.250273 | 0.235242 | 0.568008 | 0.235243 | 0.235243 | 0.266448 |
| Mean | 0.235242 | 0.235542 | 0.235243 | 0.247341 | 0.235743 | 0.235242 | 0.410957 | 0.235242 | 0.235242 | 0.239091 |
| Std | 1.41E-16 | 0.000445 | 6.97E-08 | 0.001062 | 0.002744 | 1.41E-16 | 0.081127 | 5.97E-08 | 2.39E-08 | 0.008781 |
| Rank | 2 | 6 | 5 | 9 | 7 | 1 | 10 | 4 | 3 | 8 |

**Table 24:** wilcoxon signed rank test

| mSHO vs. | SHO | DO | CMA-ES | HGS | AOA | SAO | HHO | PSO | SPO |
|----------|-----|-----|--------|-----|-----|-----|-----|-----|-----|
| Pressure vessel design problem | 3.02E-11 | 1.41E-09 | 1.72E-12 | 8.001936 | 1.07E-07 | 3.02E-11 | 3.02E-11 | 3.02E-11 | 3.16E-10 |
| Speed Reducer problem | 3.02E-11 | 8.72E-10 | 1.72E-12 | 1.14E-11 | 8.005757 | 3.02E-11 | 3.02E-11 | 1.14E-11 | 1.52E-13 |
| Tension/compression spring problem | 9.92E-11 | 1.07E-09 | 1.72E-12 | 5.89E-11 | 5.46E-11 | 3.68E-07 | 3.02E-11 | 3.02E-11 | 3.02E-11 |
| Welded beam design problem | 3.02E-11 | 7.04E-07 | 2.36E-12 | 1.21E-10 | 8.035137 | 3.02E-11 | 3.02E-11 | 8.016955 | 8.87E-10 |
| Three-bar truss engineering design problem | 3.02E-11 | 8.012732 | 1.72E-12 | 3.01E-11 | 4.12E-15 | 3.02E-11 | 3.02E-11 | 1.18E-08 | 5.46E-09 |
| Industrial refrigeration system problem | 3.02E-11 | 8.000225 | 1.72E-12 | 9.2E-05 | 8.149449 | 3.02E-11 | 3.02E-11 | 8.000691 | 3.02E-11 |
| Multi-Product batch plant problem | 3.34E-11 | 8.994102 | 3.69E-11 | 8.002499 | 1.73E-06 | 3.02E-11 | 3.34E-11 | 8.001518 | 8.46E-11 |
| Cantilever Beam Design | 3.02E-11 | 2.15E-10 | 1.72E-12 | 3.02E-11 | 3.02E-11 | 3.02E-11 | 3.02E-11 | 8.1E-10 | 3.02E-11 |
| Multidisc clutch brake problem | 3.02E-11 | 5.09E-08 | 2.36E-12 | 4.56E-11 | 1.21E-12 | 3.02E-11 | 2.93E-09 | 1.21E-12 | 8.063525 |

### 7.3 Dimic ussion

The aforementioned results show that the proposed mSHO has advanced results compared to the other metaheuristic algorithms, including SHO, DO, CMA-ES, HGS, AOA, SAO, HHO, PSO, and SPO. In addition, as optimization issues get more challenging, mSHO's effectiveness remains unchanged, demonstrating its stability and aptitude for addressing challenging search domains. This demonstrates that it is a powerful tool for addressing challenging optimization problems. The results can be summarised as follows:

· CEC'2020 test function

  – mSHO demonstrates highly competitive fitness values, ranking first for all functions except F6 and F10.
  – The proposed mSHO algorithm achieves an overall ranking of 1.
  – According to the Friedman test, the proposed mSHO exhibits the lowest value of 1.3.

· Engineering problems

  (i) Pressure vessel design problem: The optimal function value is 0.012665, attained using the mSHO algorithm.
  (ii) Speed reducer design problem: The optimal function value is 2993.634, achieved using the mSHO, HGS, AOA, and PSO algorithms.
  (iii) Tension/compression spring problem: The mSHO algorithm achieves the optimal function value of 0.01266.
  (iv) Welded beam design problem: The optimal function value is 1.724967, obtained using the mSHO algorithm.
  (v) Three-bar truss engineering design problem: The mSHO algorithm attains the optimal function value of 263.8915.
  (vi) Industrial refrigeration system problem: The mSHO algorithm reaches the optimal function value of 0.032255.
  (vii) Multi-Product batch plant problem: The mSHO algorithm yields the optimal function value of 58507.14.

(viii)  Cantilever beam problem: The optimal function value is 1.339956, achieved using the mSHO algorithm.

(ix)  Multiple disc clutch brake problem: The mSHO algorithm, in conjunction with the AOA algorithm, achieves the minimum objective function value of 0.23524.

This study is limited to the selected problems, which can be extended to experiment mSHO with machine-learning-related problems, such as feature engineering and selection, hyperparameter tuning, ensemble Learning, neural architecture Search, model Selection, and model Compression. In addition, the study is limited to single-objective optimization, which can be extended to solve multi-objective optimization problems to represent trade-offs between conflicting objectives.

## Acknowledgements

This work was supported by the Ministerio Español de Ciencia e Innovación under project number PID2020-115570GB-C22 MCIN/AEI/10.13039/501100011033 and by the Cátedra de Empresa Tecnología para las Personas (UGR-Fujitsu).

## Conflict of interest statement

The authors declared no potential conflicts of interest with respect to the research, authorship, and/or publication of this article.

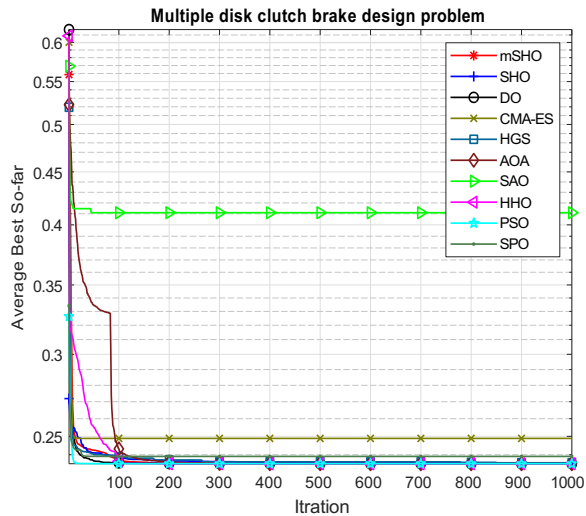

**(a)** Convergence Curve

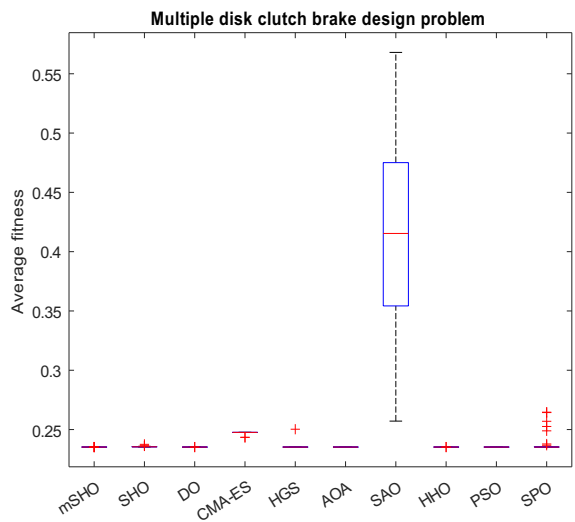

**(b)** Boxplot

**Figure 12:** Convergence curve and Boxplot for mSHO against other competitors – Multidisc clutch brake problem